%% file: main.tex
\documentclass[10pt,twocolumn]{article}
\usepackage[a4paper,margin=0.72in,columnsep=0.24in]{geometry}
\usepackage[T1]{fontenc}
\usepackage{lmodern,microtype}
\usepackage{amsmath,amssymb,booktabs,graphicx,longtable}
\usepackage{xcolor,fancyhdr,etoolbox}
\usepackage{hyperref}
\definecolor{blueink}{HTML}{235889}
\hypersetup{colorlinks,allcolors=blueink,pdfauthor={Yimin Du, Chenglin Guo, Wan Wang, Guolin Tang},pdftitle={Cognitive Decision Routing in Large Language Models: When to Think Fast, When to Think Slow}}
\makeatletter
\patchcmd{\@maketitle}{{\LARGE \@title \par}\vskip 1.5em}
  {{\LARGE \@title \par}\vskip 1em}{}{}
\patchcmd{\@maketitle}{\vskip 2em}{\vskip 1.5em}{}{}
\patchcmd{\@maketitle}{\par\vskip 1.5em}{\par\vskip .85em}{}{}
\makeatother
\title{\textbf{Cognitive Decision Routing in Large Language Models:}\\
\textbf{When to Think Fast, When to Think Slow}\thanks{Evaluation data and protocols are available at \href{https://github.com/initial-d/reasoning-state-evaluation-data}{\nolinkurl{github.com/initial-d/reasoning-state-evaluation-data}}.}}
\author{\normalsize Yimin Du \quad Chenglin Guo \quad Wan Wang \quad Guolin Tang\\
\small \texttt{sa613403@mail.ustc.edu.cn}}
\date{}
\begin{document}
\maketitle
\thispagestyle{fancy}

\begin{abstract}
\input{abstract-result.tex}
\end{abstract}

\section{Introduction}

\begin{figure*}[t]
\centering
\includegraphics[width=\textwidth]{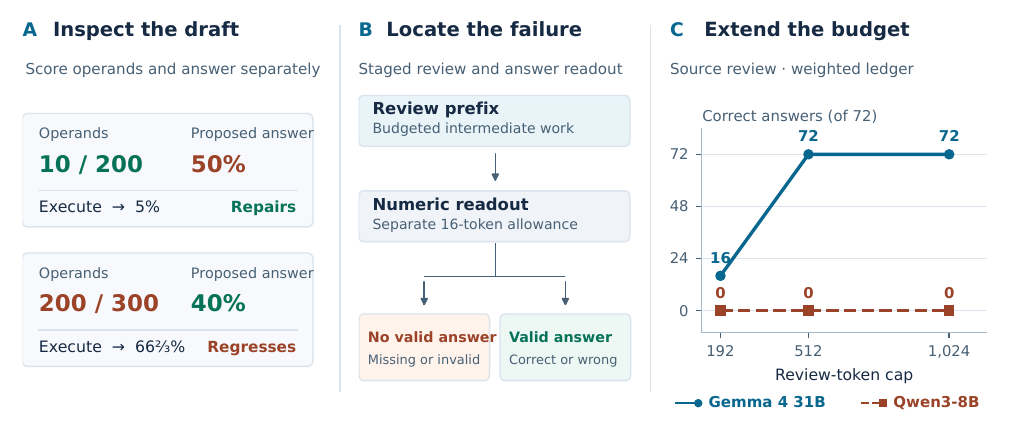}
\caption{\textbf{Reasoning-state repairability.}
(A) Two naturally generated drafts. Executing the displayed ratio
repairs the first answer and makes the second incorrect.
(B) Staged review separates intermediate work from the validity of
the answer readout.
(C) Source Review on 24 weighted-ledger questions, each with three
operand corruptions and a correct proposed answer. All 72 readouts
per point are valid. The three caps are read from shared trajectories
for each question and draft.}
\label{fig:repairability-overview}
\end{figure*}

One model draft records \(10/200\) and proposes \(50\%\).
Executing the ratio recovers the correct \(5\%\). Another draft
correctly proposes \(40\%\) while recording \(200/300\).
Executing that ratio would discard the correct answer.
Both drafts arose on conditional-probability questions.
Judging what to do with a draft requires knowing which parts
of it can be reused (Figure~\ref{fig:repairability-overview}).

Cognitive Decision Routing starts from a question about when additional
reasoning is worth using. Its query-level framework distinguishes four
signals: the strength of the relationship between the evidence and the
conclusion, whether the task crosses knowledge domains, how many
stakeholders it involves, and how uncertain or ambiguous it is.
\par
\newpage
Weakly diagnostic evidence motivates a specific prediction. It should
leave more room for deliberate reasoning to improve an answer.
Our matched evidence experiment directly tests this prediction on
exact-count questions. The observed gains from more scratch work
vary across instructions and checkpoints, and the weak-cue advantage
is inconclusive (Section~\ref{sec:cue-diagnosticity}).

Once a draft exists, routing becomes a choice about what to do with
the work already produced. An action is useful to the extent that it
can correct what is wrong while preserving what is right. We study
this property as \emph{reasoning-state repairability}. For a fixed
question, checkpoint, and budget, we ask which actions preserve a
correct answer or recover a wrong one when the process and proposed
answer are varied independently. We use a task in which source
operands can be checked exactly, and compare keeping the proposed
answer, executing the recorded calculation, reviewing the draft,
checking the source, and restarting. The four states separate the
effect of a wrong operand from the effect of a wrong answer, while
holding the original question and the draft format fixed.

Natural drafts supply a second test. We classify observable errors
against the source counts and replay the same actions for a subset
of those questions. The diagnostic labels let us ask what is
repairable under known conditions. The actions themselves see the
draft and question and receive no correctness labels. A separate
study on public arithmetic questions tests the value of retaining
an earlier calculation when its conclusion has been deliberately
changed, alongside a replay of unaltered wrong answers.

Can a draft make a problem harder for a model that can solve it from
scratch? We compare review and restart on the same questions, then
increase the allowance for both. This distinguishes errors in
reviews that finish with tokens to spare from errors that disappear
when the readout sees more of the review. A separate generation
control tests whether the model reaches the final answer at all.

The experiments make three contributions.
First, the paired draft interventions measure the distinct value of
intermediate calculations and of proposed answers.
Second, paired corruptions show when a review carries forward a
wrong operand and can turn a correct proposal into a wrong answer.
Third, the budget experiments separate a valid but incorrect repair
from a failure to finish the answer, and measure when additional tokens
change either outcome.

\newpage
\section{Related Work}

Process supervision and process verification distinguish the validity
of intermediate work from the final answer
\cite{verify,processbench}. Faithfulness studies perturb visible
traces to test their influence on answers
\cite{turpin,tracefaithfulness,answerattribution}. Error-location
experiments distinguish finding a mistake from correcting it once
localized \cite{mistakelocation}. Our question is how independently
changing an observable process error and final-answer error alters
the outcome of several actions under the same budget.

Self-refinement can repair mistakes and overwrite correct answers
\cite{selfrefine,huang,riskrevision}. Tool feedback, verbal reflection
across trials, and training for self-correction can change that balance
\cite{critic,reflexion,score,kamoi,stechly}. Chain-of-Verification
answers verification questions independently to reduce draft influence
\cite{cove}. Our Source Review keeps the draft visible while asking the
model to re-extract the operands. Restart provides a draft-free comparison.
Symbolic reasoning systems separate the construction of a program
from its execution \cite{pot,faithfulcot}. This distinction matters
for repair. Execution uses the stated operands, whereas source review
can replace them with counts extracted again from the question.
Contextual drag measures how failed attempts induce structurally
similar errors, and role relabeling shows that the presentation of a
fixed error changes whether models flag it \cite{contextdrag,roleedit}.
Early-step interventions and prefix resampling study when errors in a
reasoning trajectory persist \cite{lostbeginning,tracepropagation}.
GSM-Symbolic varies numbers or clauses in the problem
\cite{gsmsymbolic}. We hold the problem and message role fixed and vary
the correctness of individual operands and of the proposed answer.
This lets us observe which edited operands a reviewer repeats and
whether it preserves an answer that was correct before review.

Compute-optimal inference allocates effort according to problem
difficulty \cite{snell}. Budget forcing extends generation, and
prefix evaluations examine how accuracy changes after additional
reasoning \cite{s1,zhouoverthink,thinkingpast}. Studies of answer
delivery distinguish wrong completions from unfinished responses and
test cues that elicit an answer under a shared cap
\cite{competingreasoning,answerdelivery}. In our paired review task,
staged readout measures what answer the model gives with a specified
prefix. Single generation measures whether it delivers an answer
without a separate cue. The draft-free control tests whether the
same checkpoint can solve the question without the edited state.

\input{cue-diagnosticity-results.tex}
\input{final-measures.tex}
\input{repair-methods.tex}

\section{Which reasoning states can be repaired?}
\label{sec:repair-results}

\input{repair-factorial-results.tex}
\input{repair-cells.tex}
\input{repair-validity.tex}
\input{repair-natural-results.tex}
\input{final-two-task.tex}

\section{Reusing calculations on public arithmetic tasks}

\input{answer-edit-methods.tex}
\input{answer-edit-results.tex}

\input{repair-archived-evidence.tex}
\input{repair-discussion.tex}

\section{Limitations}

The conditional-count experiment tests cue diagnosticity where the
source counts determine the answer exactly. It leaves the roles
of domain crossing, stakeholder multiplicity, and uncertainty open.
The ledger requires several steps to obtain the operands, but the
intervention edits only the resulting counts or the proposed answer.
The reported effects apply to those visible draft states and the
tested actions and budgets. In the archival audit, ambiguous or
unparseable operands are classified as unknown.

The public draft edits are conditioned on a correct initial readout,
which selects problems the checkpoint could already solve. The draft
can also state the correct number in its calculation. The naturally
wrong replays cover two checkpoints and parseable integer failures.
Greedy generation gives one trajectory or completion per condition.
Results for the staged protocol depend on how the visible prefix is
read. The neutral restart preserves message roles and changes the
input length by removing the draft. The single-generation comparison
uses a different answer format and allows the model to allocate
tokens between work and the final answer. Reference answers are
used for construction and scoring. Each action runs independently
of the correctness label.

\section{Conclusion}

\input{repair-conclusion.tex}

\paragraph{Data availability.}
The ancillary materials contain the experimental prompts, model and
generation settings, synthetic inputs and reference answers, retained
raw responses and review traces, benchmark item identifiers, and
scored results. The tables can be reconstructed from the archived
question-level records and the scoring and analysis procedures
specified in the paper and accompanying protocols.

\bibliographystyle{unsrt}
{\fontsize{8.5}{9.7}\selectfont
\let\originalbibliography\thebibliography
\renewcommand{\thebibliography}[1]{%
  \originalbibliography{#1}%
  \setlength{\itemsep}{0pt plus .4pt}%
  \setlength{\parsep}{0pt}%
}
\bibliography{references}}
\input{repair-supplement.tex}
\end{document}

%% file: abstract-result.tex
When does more reasoning improve a model's answer? In matched
probability questions, weak evidence does not reliably predict
greater gains from longer reasoning. Once a draft exists, we ask
when it helps a reviewer and when it makes a solvable problem
harder. We independently edit the numerator, denominator, and
proposed answer, vary whether a wrong answer agrees with the
displayed calculation, and compare five actions under fixed budgets.
Across 504 natural drafts, denominator errors dominate the wrong
numeric answers, and executing the recorded operands overwrites nine
of 13 right answers backed by a wrong pair. On a multistage ledger,
Gemma 4 31B and Qwen3-8B produce valid answers yet often repeat
corrupted operands despite being asked to re-extract them. The
predefined comparison finds no reliable advantage of source over
generic review on corrupted relative to exact states. Longer review
prefixes let 31B preserve all 72 correct proposals with corrupted
operands at 512 tokens. On questions that Qwen3-14B solves from
scratch, Source Review finishes within 40 tokens and loses 61
of 63 correct proposals. Generic Review loses 7 at a 1,024-token
cap on the same drafts.
On separate count questions, single generation produces a large source-review
advantage entirely through answer completion. These experiments
distinguish information available for reuse, errors introduced or
repaired during review, and whether an answer is delivered within
budget.

%% file: cue-diagnosticity-results.tex
\section{Does weaker evidence justify more reasoning?}
\label{sec:cue-diagnosticity}

Work on representativeness motivates varying how informative a cue is
while holding its description fixed \cite{prediction}. Metareasoning and
resource-rational analysis motivate measuring the accuracy gained from
additional computation \cite{metareasoning,resourcerational}.

We test whether a cue with little diagnostic value predicts a larger
accuracy gain from additional work. Each matched group contains a
nondiagnostic world, where \(P(Y=1\mid X=1)=P(Y=1)\), and two worlds
where the cue moves the target up or down. The change in the conditional
target measures diagnosticity for the observed attribute. Mutual
information instead averages over all values of the attribute and
outcome. Let \(G_w\) be the change in exact accuracy between two
policies or caps in world \(w\), averaged over the two typography
conditions. We estimate
\[
  I_{\mathrm{cue}} =
  G_{\mathrm{null}} -
  \tfrac{1}{2}(G_{\mathrm{aligned}}+G_{\mathrm{reverse}}).
\]
If a weak cue gives more room for the selected intervention to help,
this interaction should be positive. The primary readout parser
counts invalid answers as incorrect. The analysis reuses the paired
confirmation and reserve responses, resamples complete groups, and
adjusts across 15 exploratory model-by-contrast comparisons.

\begin{table}[t]
\centering
\small
\setlength{\tabcolsep}{3pt}
\caption{Effect of raising the scratch cap from 24 to 160 on
confirmation tasks. Gains and interactions are in percentage points.
Intervals adjust for all 15 cue-diagnosticity comparisons.}
\label{tab:cue-diagnosticity}
\begin{tabular}{@{}llrrr@{}}
\toprule
Model & Policy & Null & Diagnostic & \(I_{\mathrm{cue}}\) [CI] \\
\midrule
Qwen & Free & $9.4$ & $30.2$ & $-20.8$ [$-49.5$, $8.2$] \\
 & Concise & $35.4$ & $33.9$ & $1.6$ [$-12.5$, $16.1$] \\
 & Structured & $13.5$ & $15.1$ & $-1.6$ [$-15.6$, $12.0$] \\
E2B & Free & $10.4$ & $9.4$ & $1.0$ [$-13.0$, $16.1$] \\
 & Concise & $14.6$ & $13.5$ & $1.0$ [$-21.4$, $22.7$] \\
 & Structured & $-27.1$ & $-24.0$ & $-3.1$ [$-11.5$, $4.7$] \\
E4B & Free & $-4.2$ & $7.3$ & $-11.5$ [$-35.9$, $9.7$] \\
 & Concise & $0.0$ & $-0.5$ & $0.5$ [$-2.1$, $4.7$] \\
 & Structured & $0.0$ & $0.0$ & $0.0$ [$-22.9$, $22.9$] \\
\bottomrule
\end{tabular}
\end{table}

None of the 15 adjusted intervals excludes zero. Within a fixed
instruction, the budget increase has mixed effects
(Table~\ref{tab:cue-diagnosticity}). At the same 80-token cap, switching from
concise to structured scratch yields positive point estimates in
every diagnosticity stratum, but five of the six confirmation or
reserve interactions favor diagnostic cues. The weak-cue condition
therefore lacks a reliable advantage for either intervention.
These are exact-count problems with visible evidence, and the
intervals allow meaningful interactions in either direction.

When a resampled contrast is constant, we report a conservative
bound instead of a zero-width interval. The table uses an exact
departure-probability bound conditional on the first of 48 groups,
with the \(0.05/15\) familywise allocation and the known contrast
range of \([-2,2]\). The mixed effects motivate measuring what a
particular draft retains or loses after reasoning has begun.

\begin{table}[t]
\centering\small\setlength{\tabcolsep}{3pt}
\caption{At 80 scratch tokens, gain from structured over concise
work, by cue type. Values are points. The final column is the
adjusted interval for the weak-minus-diagnostic gain.}
\label{tab:fixed-cue}
\begin{tabular}{@{}llrrr@{}}
\toprule
Cohort & Model & Weak & Diagnostic & Interaction [CI] \\
\midrule
Main & Qwen & 1.0 & 4.7 & \(-3.6\ [-22.4,15.6]\) \\
 & E2B & 11.5 & 19.3 & \(-7.8\ [-25.3,8.2]\) \\
 & E4B & 16.7 & 15.1 & \(+1.6\ [-17.5,19.3]\) \\
Reserve & Qwen & 8.5 & 16.2 & \(-7.7\ [-22.5,7.7]\) \\
 & E2B & 22.5 & 31.7 & \(-9.2\ [-25.4,5.6]\) \\
 & E4B & 4.2 & 5.6 & \(-1.4\ [-18.2,14.8]\) \\
\bottomrule
\end{tabular}
\end{table}

The fixed-cap comparison (Table~\ref{tab:fixed-cue}) varies review
structure while holding the nominal generation cap fixed. Its
improvements do not establish a benefit of increasing the cap.
We treat it as a separate check on whether a weak cue identifies
greater scope for an intervention to help.

%% file: final-measures.tex
\section{Measuring repairability}
\label{sec:repairability-definition}

A reasoning state is the visible intermediate information and, when
present, a proposed answer for a question. We define its
\emph{repairability} relative to an action, checkpoint and generation
protocol at a stated budget. Given a state class, we measure the chance
that the action returns the correct answer when the initial answer is
wrong, and the chance that it loses correctness when the initial answer
is right. References determine the state class for diagnostic analysis.
The action sees the question and the displayed draft.

Let \(Y_0\) indicate a correct proposed answer, \(V\) a valid final
answer, and \(Y\) a correct final answer, with \(Y=0\) when \(V=0\).
We report the repair rate \(P(Y=1\mid Y_0=0)\), regression rate
\(P(Y=0\mid Y_0=1)\), completion \(P(V=1)\), and completed-answer
accuracy \(P(Y=1\mid V=1)\). A conditional rate is left undefined for
an empty denominator. Invalid initial answers are a separate archival
stratum. Overall correctness decomposes into completion and accuracy
conditional on completion. This decomposition is descriptive because
the set of completed responses can change under an intervention.
Mean realized review and readout lengths measure generation cost, and
review input lengths describe prompt size. Tokens are checkpoint-specific
cost measures. Runtime and hardware energy are outside these measurements.

Operand correctness is determined by source arithmetic. It records
whether the state contains information that can be reused exactly.
A wrong pair can still give the correct ratio, so we score pair correctness
and answer correctness separately. In the two-task study, an explicit
working pair at the start of the review supplies a third observation.
We distinguish a reviewer that states the exact pair, repeats the
edited pair, states another pair, or leaves the pair unscorable.
These are observable behaviors under paired prompts and draft edits.
A printed pair alone cannot establish how the model internally computed
the answer.

%% file: repair-methods.tex
\section{A paired intervention on the draft}
\label{sec:repair-design}

\paragraph{Task.}
Each question describes a finite audit with a binary outcome and an
observed attribute. To find the requested probability, a solver selects
the count \(n\) with both properties and the count \(d\) with the observed
attribute, then computes \(100n/d\). We reuse a generator that renders
the same sufficient counts as a table or a set of statements. The input
is identical across all interventions on a question.

\paragraph{Process and answer.}
The draft represents the process through its two source operands.
A correct process supplies the exact pair \((n,d)\). A wrong process
supplies \((d-n,d)\), taking the count for the opposite outcome from
the observed slice. The proposed answer is a separate line, independently
set to the correct value or a seeded change of 5 or 10 percentage points.
The wrong answer differs from both the target and the value implied by
the wrong pair. The four states share wording, message roles, and
tokenized input length within each review action. We filter for those
controls before inference. The user question, assistant draft, and
user action instruction appear in that order. This design addresses
operand selection and answer commitment in a short, verifiable process. A correct operand
pair gives evidence from which a solver can compute the answer, and
the wrong-process/wrong-answer cell contains two discordant values.

\paragraph{Actions.}
Keep returns the proposed answer. Execute computes the ratio from the
visible operands, falling back to Keep when there is no valid pair.
Generic Review asks the model to check the calculation. Source Review
asks it to re-extract the matching counts from the original audit.
Restart replaces the draft with a neutral assistant message and gives
the same question to the model. Restart is evaluated once per question
and cap and acts as a shared baseline across the four states.
No action receives a reference answer, a correctness label, or an
error location as an additional input. A correct proposed answer is
visible only when called for by the intervention or emitted by the
initial model. Exact operands can reveal the result by arithmetic.

\paragraph{Budgets and cohorts.}
Both Gemma 4 instruction-tuned checkpoints generate greedily in
non-thinking mode. The staged protocol produces a single review up to
320 tokens and reads a numerical answer from its 64- and 320-token prefixes, with up to 16
tokens for each readout. The E4B screen uses 48 questions, one per
archived scenario group, sampled without reading earlier outcomes.
For each selected question we also replay its archived draft. The analysis
plan specifies an independent 72-question Gemma 4 31B replication if
the screen's primary interaction is at least 10 points in absolute
value and at least 85\% of paired bootstrap estimates share its sign.
The replication includes a normal-generation condition in which the
model writes its work and final answer in one turn. The two total caps
in that condition are 80 and 336 tokens. A missing or unparsable
numerical answer counts as incorrect.

\paragraph{Estimand.}
For each question and budget let \(D_{p,a}\) be the correctness
difference between Source Review and Generic Review when the displayed
process and answer have statuses \(p\) and \(a\). The primary contrast
at 320 tokens is
\begin{equation}
\label{eq:state-interaction}
\tau_{320}=\tfrac12\sum_{a}
 \mathbb E[D_{\mathrm{wrong},a}-D_{\mathrm{correct},a}].
\end{equation}
We resample questions jointly across cells and report two-sided 95\%
percentile intervals using 20{,}000 draws. The interaction at 64 tokens,
the difference between budgets, the interaction with answer status, and
all action-level comparisons are secondary. We count repairs and
regressions relative to Keep.

%% file: repair-factorial-results.tex
\subsection{The same calculation can repair or corrupt an answer}

The two deterministic actions make opposite commitments. Keep
preserves the proposed answer irrespective of the operands. Execute
repairs every wrong answer backed by exact operands, but overwrites
every correct answer backed by the wrong pair (Table~\ref{tab:repair-cells}).
These are checks on the construction. In natural drafts the
implication need not hold: a wrong pair can imply the right ratio.

At 64 review tokens on E4B, Generic and Source Review correct 47
and 48 of 48 wrong answers backed by exact operands. When the
operands and answer are both wrong, they correct only 5 and 6.
Increasing the cap to 320 improves these latter counts to 24 and
31, but reduces the former to 41 and 44. Even with a fully
correct state, each review regresses on three questions at 320
tokens, where each preserved all 48 answers at 64. Extra review
helps most when the draft's operands need replacing.

\paragraph{An exploratory interaction on E4B.}
At 320 tokens the source instruction adds 16.7 points over generic
review with wrong operands, averaging over the two answer states,
and 3.1 points with exact operands. The primary difference is
\(+13.5\) points (95\% CI \([1.0,28.1]\)). The 64-token estimate
is \(+1.0\) (\([-2.1,5.2]\)); the change with budget is \(+12.5\)
(\([-1.0,27.1]\)). The process penalty, averaged over both answer
states, falls from 87.5 to 30.2 points under Generic Review and from
86.5 to 16.7 under Source Review. The E4B interaction meets the
pre-specified threshold for a separate 31B test.

The proposed answer itself can be useful even with wrong operands.
At 320 tokens, E4B Source Review is correct on 42 versus 31 questions
with wrong operands, depending on whether the proposed answer is
correct. With exact operands, the corresponding counts are 45 and
44. The exploratory difference between those answer effects is
20.8 points (95\% CI \([10.4,33.3]\)). The last line of a damaged
draft can retain information that its calculation has lost.

\begin{figure}[t]
\centering
\includegraphics[width=.98\columnwidth]{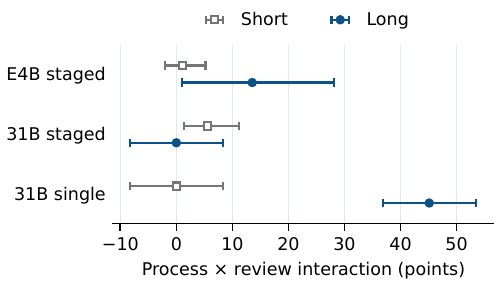}
\caption{The process-by-review interaction in
Equation~\ref{eq:state-interaction} with paired 95\% intervals. Short and long are 64 and 320 review tokens plus a
16-token answer request in the staged protocol, and 80 and 336
total tokens in single generation. The staged and single-generation
31B results use the same 72 questions.}
\label{fig:repair-interaction}
\end{figure}

\paragraph{On 31B, the answer protocol changes the result.}
The pre-specified staged replication gives an interaction of zero
at 320 tokens. Both review actions, and Restart, solve all 72 questions
in every cell. We report a conservative 95\% interval of
\([-8.3,8.3]\) points from the exact departure bound conditional on
the first question and the score support \([-2,2]\).
At 64 tokens, the staged interaction is \(+5.6\) points (95\% CI
\([1.4,11.1]\)). Source Review solves every question, while Generic
Review misses eight wrong-operand cases across the two answer states.
The interaction therefore decreases with the cap on 31B. The E4B
screen estimates an increase (Figure~\ref{fig:repair-interaction}).

In the single-generation control, no review or restart produces
a valid final line at 80 tokens (Table~\ref{tab:repair-validity}). At 336 tokens the interaction is
\(+45.1\) points (95\% CI \([36.8,53.5]\)). Source Review gives
a valid, correct answer in every cell. With wrong operands Generic
Review finishes with a valid answer on 26 of 72 questions when the
proposed answer is correct and 50 of 72 when the proposed answer
is wrong. With exact operands it finishes on 71 and 70 questions,
respectively. Every valid Generic Review answer is correct and
every invalid one reaches the cap. The accuracy interaction is
therefore also an interaction in finishing with a valid answer.
The staged and single-generation protocols act differently even at
the same total cap. The comparison varies the final-answer format
and whether the model gets a separate cue to stop reviewing and
commit to an answer.

%% file: repair-cells.tex
\begin{table*}[t]
\centering\small
\setlength{\tabcolsep}{3.5pt}
\caption{All paired arm outcomes. Counts are correct answers out of the stated sample. Process and answer columns give the correctness of the displayed operands and proposed answer. The short and long caps are 64 and 320 review tokens plus a 16-token readout, or 80 and 336 total tokens for single generation. Restart is shared across the four cells. Gen. denotes Generic Review, and Src. denotes Source Review.}
\label{tab:repair-cells}
\resizebox{.98\linewidth}{!}{%
\begin{tabular}{@{}llrrrrr@{\hspace{1.05em}}rrrrr@{}}
\toprule
& & \multicolumn{5}{c}{Short cap} & \multicolumn{5}{c}{Long cap} \\
\cmidrule(lr){3-7}\cmidrule(lr){8-12}
Checkpoint and protocol & Process / Answer & Keep & Exec. & Gen. & Src. & Restart & Keep & Exec. & Gen. & Src. & Restart \\
\midrule
E4B, staged ($n=48$) & Exact / Correct & 48 & 48 & 48 & 48 & 7 & 48 & 48 & 45 & 45 & 20 \\
 & Exact / Wrong & 0 & 48 & 47 & 48 & 7 & 0 & 48 & 41 & 44 & 20 \\
 & Wrong / Correct & 48 & 0 & 6 & 7 & 7 & 48 & 0 & 33 & 42 & 20 \\
 & Wrong / Wrong & 0 & 0 & 5 & 6 & 7 & 0 & 0 & 24 & 31 & 20 \\
\addlinespace[2pt]
31B, staged ($n=72$) & Exact / Correct & 72 & 72 & 72 & 72 & 68 & 72 & 72 & 72 & 72 & 72 \\
 & Exact / Wrong & 0 & 72 & 72 & 72 & 68 & 0 & 72 & 72 & 72 & 72 \\
 & Wrong / Correct & 72 & 0 & 67 & 72 & 68 & 72 & 0 & 72 & 72 & 72 \\
 & Wrong / Wrong & 0 & 0 & 69 & 72 & 68 & 0 & 0 & 72 & 72 & 72 \\
\addlinespace[2pt]
31B, single ($n=72$) & Exact / Correct & 72 & 72 & 0 & 0 & 0 & 72 & 72 & 71 & 72 & 65 \\
 & Exact / Wrong & 0 & 72 & 0 & 0 & 0 & 0 & 72 & 70 & 72 & 65 \\
 & Wrong / Correct & 72 & 0 & 0 & 0 & 0 & 72 & 0 & 26 & 72 & 65 \\
 & Wrong / Wrong & 0 & 0 & 0 & 0 & 0 & 0 & 0 & 50 & 72 & 65 \\
\bottomrule
\end{tabular}}
\end{table*}

%% file: repair-validity.tex
\begin{table}[t]
\centering\small
\caption{Valid numerical answers (\%). Review validity is shown as a range across the four factorial cells. Restart has one rate per budget. Keep and Execute return a valid number on every manufactured draft.}
\label{tab:repair-validity}
\setlength{\tabcolsep}{3.5pt}
\begin{tabular}{@{}llrrr@{}}
\toprule
Protocol & Cap & Generic & Source & Restart \\
\midrule
E4B, staged & Short & 100.0 & 97.9--100.0 & 52.1 \\
 & Long & 97.9 & 93.8--100.0 & 66.7 \\
\addlinespace[2pt]
31B, staged & Short & 100.0 & 100.0 & 100.0 \\
 & Long & 100.0 & 100.0 & 100.0 \\
\addlinespace[2pt]
31B, single & Short & 0.0 & 0.0 & 0.0 \\
 & Long & 36.1--98.6 & 100.0 & 90.3 \\
\bottomrule
\end{tabular}
\end{table}

%% file: repair-natural-results.tex
\section{Diagnosing naturally generated drafts}
\label{sec:repair-natural}

We audit 504 archived E4B drafts from 84 scenario groups. We independently
extract the reference operands from the source text and verify the archived
operands and answer on every question. A draft is scorable for process
correctness if it contains one unambiguous labeled operand pair.
We distinguish an incorrect numerator, an incorrect
denominator, errors in both counts, and a correct pair with a wrong
answer. We retain ambiguous or invalid pairs and unparseable answers as
separate categories. The audit checks whether the final number agrees
with the ratio from the emitted counts. A wrong pair can give the right
ratio, so operand correctness and outcome correctness remain separate.

The archive includes outcomes of Generic and Source Review at 64 and
160 tokens. Those replays have their own original prompts. The new
48-question replay uses the same review prompts and caps as the
factorial experiment and permits a within-question comparison with the
artificial state of the same verified correctness. Naturally occurring
errors can change both operands, arithmetic, and how the draft is
written. The paired comparison is descriptive because these aspects
are not held fixed. Gold labels enter experimental construction and
offline scoring. They do not enter action prompts or select actions.

\subsection{Errors in natural drafts often involve the denominator}

Of 504 archived E4B drafts, 154 have a correct numeric answer,
239 give a wrong number, and 111 have an invalid answer readout.
Among the 239 wrong numeric answers, 136 have only a wrong denominator,
90 have two wrong operands, four have only a wrong numerator, and two
have exact operands. Seven have an unscorable operand pair.
In 86 cases the wrong answer matches the emitted ratio. A further
six drafts with a wrong pair have operands whose ratio equals the
reference answer. These categories distinguish wrong source selection
from a failure to commit to the value already implied by the draft.

\begin{table}[t]
\centering
\small
\setlength{\tabcolsep}{3.4pt}
\caption{Archived E4B drafts with a wrong numeric answer. Cells count
correct answers after each action. The final column gives stratum sizes.}
\label{tab:natural-types}
\begin{tabular}{@{}lrrrrrr@{}}
\toprule
& & \multicolumn{2}{c}{64 tokens} & \multicolumn{2}{c}{160 tokens} & \\
\cmidrule(lr){3-4}\cmidrule(lr){5-6}
Operand status & Exec. & Gen. & Src. & Gen. & Src. & \(n\) \\
\midrule
Exact & 2 & 2 & 2 & 2 & 1 & 2 \\
Wrong numerator & 0 & 4 & 1 & 4 & 2 & 4 \\
Wrong denominator & 0 & 42 & 0 & 60 & 45 & 136 \\
Both wrong & 2 & 33 & 4 & 34 & 12 & 90 \\
Unscorable & 0 & 1 & 0 & 1 & 6 & 7 \\
\bottomrule
\end{tabular}
\end{table}

Execution fixes the two observed wrong answers with exact operands
and generally inherits wrong counts (Table~\ref{tab:natural-types}).
Conversely, 13 drafts give the right answer with a wrong pair. Executing
their reported ratio preserves only four of those answers. The
numerator-only and answer-only strata are small. The largest natural
strata involve the denominator, which the controlled numerator edit
holds fixed. For denominator-only errors, Source Review repairs
0 of 136 at 64 tokens and 45 at 160. Generic Review repairs 42 and 60,
respectively.

\paragraph{Matching questions does not match the errors.}
Of the 48 replayed drafts, 26 have a scorable wrong pair and a wrong
numeric answer. At 320 tokens, Generic Review repairs 7 of 26
natural drafts and 14 of 26 matched artificial drafts. Source Review
repairs 16 and 18, respectively. The paired natural-minus-artificial
gaps are \(-26.9\) points (95\% CI \([-50.0,0.0]\)) for Generic
and \(-7.7\) (\([-30.8,15.4]\)) for Source Review. Both reviewers
repair none of these natural drafts at 64 tokens and two artificial
drafts each. On the 10 pairs with exact natural operands and a
correct answer, all actions except Restart preserve all 10 answers
at 320 tokens. The comparison pairs questions and correctness
labels. It allows the location and severity of the error, its
surface form, and the coherence of the draft to differ.

%% file: final-two-task.tex
\section{Repairing denominator and joint errors}
\label{sec:two-task}

\paragraph{Tasks and cohort.}
We compare two fresh sets of 24 questions on Gemma 4 31B and Qwen3-8B.
The first requires choosing counts for a conditional percentage.
The second uses a 15-row weighted ledger. The solver classifies rows,
subtracts the smaller of each debit and its cap, applies a row weight,
then sums the eligible and verified amounts before dividing. The
questions share exact integer-percentage output. The ledger has more
upstream operations and makes each operand independently checkable.
The checkpoints receive identical questions. Selection uses a fixed
generator seed, distinct ratios and matched tokenized prompt lengths
within each reviewer on both checkpoints.

\paragraph{Location and coherence.}
For each question we construct exact, wrong-numerator,
wrong-denominator and jointly wrong pairs. The denominator distractor
includes otherwise ineligible rows. The numerator distractor selects the
alternate outcome or the reserve ledger. With each erroneous pair we
cross a correct proposal, the wrong percentage that the pair implies,
and a conflicting wrong percentage. The conflicting values are a cyclic
permutation of the three coherent wrong percentages from the same
question. This matches their multiset across the corrupted states.
Within a single location the two proposals may differ in distance from
the reference. Exact pairs receive a correct or wrong proposal. This
gives eleven states per question.

\paragraph{Action and mechanism intervention.}
The five actions are Keep, Execute, Generic Review, Source Review, and
Restart. The reviewer is asked to state its working counts first under
the same format in both review arms. Generic Review checks the draft
calculation and Source Review explicitly re-extracts the operands from
the question. Replacing one draft field at a time permits comparison
of which values the reviews carry into their written working pair.
Scoring first-line pairs is conservative. Nonmatching formatting,
inadmissible counts and an absent pair count as unscorable.
Each staged readout follows a 64- or 192-token review prefix and has
16 answer tokens. Protocols, cohort sizes and tests were fixed before
scored inference.

\paragraph{Predefined comparison.}
With the proposed answer held correct, we test whether the source
instruction's advantage on corrupted pairs exceeds its advantage on
exact pairs on the ledger at 192 review tokens. We average the
three errors within each question before resampling or sign flipping.
We report 95\% paired intervals, 100,000 two-sided sign draws under
sign exchangeability, and Holm adjustment across the two checkpoints.
The count-task, coherence, location, length and trace analyses
are descriptive. For a constant resampling contrast we condition on
the first question and use an exact upper bound on a departure among
the remaining questions. Multiplying that probability bound by the
largest departure on the contrast's support gives a conservative
interval of nonzero width.

\input{final-two-task-results.tex}
\input{ledger-budget-results.tex}

%% file: final-two-task-results.tex
\subsection{Operand reuse and answer completion}
\label{sec:new-outcomes}

\input{final-two-task-primary.tex}

\input{final-two-task-mechanism.tex}

On the weighted ledger, Qwen3-8B preserves all 24 correct proposals
with exact operands under both reviews at 192 tokens. Holding the
proposal correct while changing an operand gives one correct answer in
72 responses under Generic Review and zero under Source Review.
All 144 responses in this comparison contain a valid final percentage
(Table~\ref{tab:mechanism-main}).
In the corresponding source-review prefixes, all 72 working pairs
repeat the edited operands. The generic reviewer repeats the edited
pair in 69 of 72. The action that explicitly requests source extraction
can therefore complete the review and still reproduce the corrupted
working state. This failure persists in the longer staged condition
and differs from failure to produce a final line in single generation.

Gemma 4 31B reaches 72/72 with both reviewers on the simple count
task, but only 12/72 under Generic and 11/72 under Source Review
on the corresponding ledger states. With exact ledger operands
and a correct proposal, the same actions give 23/24 and 24/24.
Restart solves two ledger questions on Gemma and none on Qwen at
the long cap. The ledger exposes room for repair and retains a
benefit for correct intermediate work. The accuracy losses in
Table~\ref{tab:mechanism-main} occur despite full answer completion.

\paragraph{An advantage for one edit can hide a loss on another.}
Holding the proposal correct on the ledger, 31B Source versus Generic
Review preserves 8 versus 4 answers with wrong numerators at the
long cap. The pairs are 1 versus 6 for wrong denominators and
2 versus 2 for joint errors. From 64 to 192 review tokens the source
reviewer increases its numerator-only successes from 3 to 8 and
its denominator-only successes from 0 to 1. The generic reviewer
improves from 2 to 4 and 3 to 6 on those two strata.
On the simpler count questions, Qwen repeats the wrong pair in
11 of 24 Generic reviews and 9 of 24 Source reviews when only the
denominator changes and the proposal is correct. These differences locate
the sensitivity in an observable part of the input, and the
location-specific comparisons remain descriptive.

On the ledger both checkpoints exhibit copying of the edited pair.
For 31B the source instruction increases exact repetition of the
wrong pair from 24 to 40 of 72, and only two generic and zero source
reviews begin with the exact pair. Some final answers are correct
even without an exact first-line pair. The observation rules out
explaining the aggregate accuracy difference solely through answer
completion. It does not identify a causal pathway from the printed
line to the final answer.

\paragraph{The prespecified interaction and coherence test.}
The primary source-by-corruption interaction is \(-5.6\) points
on 31B and \(-1.4\) on Qwen (Holm-adjusted \(p=.867\) and \(1.000\)).
We find no consistent source advantage across the checkpoints
(Table~\ref{tab:two-task-primary}). Coherent and conflicting wrong
proposals also have similar outcomes on the ledger. On 31B at
192 tokens the coherent-minus-conflicting effect is \(-1.4\) points
for generic and \(+2.8\) for source review. On Qwen both are zero.
The paired uncertainty and near-floor recovery on Qwen limit
what can be inferred about coherence. Full action outcomes, traces,
conditional rates, and realized lengths appear in
Appendix~\ref{app:final-complete}.

%% file: final-two-task-primary.tex
\begin{table*}[t]
\centering\small
\caption{Difference in Source versus Generic accuracy between corrupt and exact operand states
with the proposal held correct. Estimates are in points with paired 95\% intervals.
Each row uses 24 questions. The final column applies Holm adjustment to the two
192-token ledger tests.}
\label{tab:two-task-primary}
\begin{tabular}{@{}llrrr@{}}
\toprule
Checkpoint & Task & 64 tokens & 192 tokens & Holm \(p\) \\
\midrule
Gemma 4 31B & Counts & \(+0.0\ [-24.4,24.4]\) & \(+0.0\ [-24.4,24.4]\) & -- \\
Gemma 4 31B & Ledger & \(-9.7\ [-27.8,4.2]\) & \(-5.6\ [-15.3,4.2]\) & 0.867 \\
Qwen3-8B & Counts & \(-6.9\ [-18.1,2.8]\) & \(-6.9\ [-18.1,2.8]\) & -- \\
Qwen3-8B & Ledger & \(-2.8\ [-6.9,0.0]\) & \(-1.4\ [-4.2,0.0]\) & 1.000 \\
\bottomrule
\end{tabular}
\end{table*}

%% file: final-two-task-mechanism.tex
\begin{table*}[t]
\centering\small
\caption{Wrong displayed operands and correct proposed answer at 192
review tokens. Each row pools 72 responses from 24 questions with three
operand corruptions each. Exact, repeated and unscorable refer to the
first written pair. Other parseable pairs make up the remaining cases.}
\label{tab:mechanism-main}
\begin{tabular}{@{}lllrrrrr@{}}
\toprule
Checkpoint & Task & Review & Completed & Correct & Exact pair & Repeated edit & Unscorable \\
\midrule
Gemma 4 31B & Counts & Generic & 72 & 72 & 72 & 0 & 0 \\
Gemma 4 31B & Counts & Source & 72 & 72 & 72 & 0 & 0 \\
Gemma 4 31B & Ledger & Generic & 72 & 12 & 2 & 24 & 0 \\
Gemma 4 31B & Ledger & Source & 72 & 11 & 0 & 40 & 0 \\
Qwen3-8B & Counts & Generic & 72 & 58 & 57 & 11 & 1 \\
Qwen3-8B & Counts & Source & 72 & 56 & 36 & 9 & 23 \\
Qwen3-8B & Ledger & Generic & 72 & 1 & 0 & 69 & 0 \\
Qwen3-8B & Ledger & Source & 72 & 0 & 0 & 72 & 0 \\
\bottomrule
\end{tabular}
\end{table*}

%% file: ledger-budget-results.tex
\subsection{Final budget sensitivity on weighted ledgers}
\label{sec:ledger-budget}

We reuse the same 24 ledgers, eleven draft states and two checkpoints.
With the prompts held fixed, we extend the review allowance to
512 and 1,024 tokens and score prefixes of the same review at 192,
512 and 1,024. Each prefix has a separate 16-token answer readout.
The pre-specified descriptive comparison averages the changes in
answer correctness across the three corrupted operand states with a
correct proposal within each question. The interval resamples
questions. We also record actual review length and the written first
pair. A longer prefix can preserve an erroneous first line while
adding a valid correction later.

\input{ledger-budget-table.tex}

The larger caps have little effect on Qwen3-8B in the main ledger
stratum. All readouts are valid at every cap. Source Review keeps
zero of the 72 correct proposals and writes the edited pair in all
72 initial lines. All 72 source-review trajectories are
37--43 tokens long (mean 38.5), so the evaluation never truncates
them at any of the three budgets. Generic Review preserves 1, 2
and 1 correct proposals as the cap rises. Only two of its 72
trajectories extend beyond 192 tokens. At 1,024 tokens neither
review arm hits the review cap or the separate 16-token answer
cap. The persistent wrong answers occur in completed responses
below both caps (Table~\ref{tab:ledger-budget}).

\paragraph{Additional review recovers the result on 31B.}
On 31B, all 72 complete review trajectories in each arm extend
beyond 192 tokens. The 192-token condition therefore supplies a
truncated review prefix to the separate answer request. Generic Review rises from 9 to 72 correct answers
at 512, and Source Review from 16 to 72. The paired changes are
\(+87.5\) and \(+77.8\) points, respectively. All readouts at the
three caps are valid. The same first lines still reproduce the
edited operands in 22 generic and 42 source reviews at 512.
The longer prefix can therefore support a correct answer even
when its first line repeats the wrong operands. At 1,024 both
reviews answer all 264 ledger states correctly. The last generic
errors at 512 occur in a coherent wrong state (two cases) and a
conflicting wrong state (one case).

\paragraph{Restart separates short reviews from task difficulty.}
At 1,024 tokens, Qwen answers 15 of 24 ledgers correctly when
restarted, with 23 valid readouts. At 192 tokens it has one valid
restart readout and zero correct answers. In contrast, the long
reviews of corrupted states all deliver valid percentages but
rarely retain the correct proposal. On the 15 questions Qwen solves
from scratch, Source Review repeats the edited pair and answers
incorrectly in all three corrupted states. Raising a cap only extends
the prefix if the review reaches it. On 31B the extended prefix
corrects the answer even when an early line is wrong.
No final answer in the main stratum reaches the 16-token answer
cap. Restart has its own readout behavior and is reported separately.

\paragraph{Matched-prefix interpretation.}
Against the original 192-token run, the 192-token replay changes
correctness in 9 of 264 generic and 10 of 264 source responses
on 31B, across the complete eleven-state design. There are no
correctness changes on Qwen. The differences in
Table~\ref{tab:ledger-budget} use the replayed prefixes from the same
long trajectory. The change between runs is reported separately
from the effect of reading a longer prefix. The full cells and
Restart controls are in Appendix~\ref{app:ledger-budget}.

\paragraph{A larger Qwen checkpoint reproduces the short Source Review failure.}
We replay the correct-proposal states on Qwen3-14B with the same
questions, prompts and caps (Table~\ref{tab:ledger-qwen14b}).
At 1,024 tokens, Restart solves 21 of 24 questions. On those 21,
Source Review preserves only 2 of 63 correct proposals with corrupted
operands. All 63 answers are valid, and all 61 wrong answers follow
reviews that begin with the edited pair. These reviews finish within
37--40 tokens at every cap. Generic Review at 1,024 preserves 56 of 63
on the same drafts. On the 12 questions both Qwen checkpoints solve
from scratch, Source Review preserves 0 of 36 proposals on 8B and
2 of 36 on 14B. The short-review failure thus persists across these
two sizes under the source instruction, while 14B can use the larger
allowance under the generic instruction.

%% file: ledger-budget-table.tex
\begin{table*}[t]
\centering\small
\caption{Final ledger sensitivity on wrong operands with a correct proposal.
Each row pools 72 responses from 24 questions, with equal weight for the
three operand locations. Repeated refers to the first written pair.
The difference from 192 reports points and a paired 95\% interval.}
\label{tab:ledger-budget}
\begin{tabular}{@{}lrlrrrrr@{}}
\toprule
Checkpoint & Cap & Review & Correct & Valid & Repeated & Mean review tokens & Change from 192 \\
\midrule
Gemma 4 31B & 192 & Generic & 9 & 72 & 22 & 192.0 & -- \\
Gemma 4 31B & 192 & Source & 16 & 72 & 42 & 192.0 & -- \\
Gemma 4 31B & 512 & Generic & 72 & 72 & 22 & 484.1 & \(+87.5\ [79.2,94.4]\) \\
Gemma 4 31B & 512 & Source & 72 & 72 & 42 & 425.4 & \(+77.8\ [69.4,86.1]\) \\
Gemma 4 31B & 1024 & Generic & 72 & 72 & 22 & 544.2 & \(+87.5\ [79.2,94.4]\) \\
Gemma 4 31B & 1024 & Source & 72 & 72 & 42 & 427.7 & \(+77.8\ [69.4,86.1]\) \\
Qwen3-8B & 192 & Generic & 1 & 72 & 69 & 53.6 & -- \\
Qwen3-8B & 192 & Source & 0 & 72 & 72 & 38.5 & -- \\
Qwen3-8B & 512 & Generic & 2 & 72 & 69 & 62.5 & \(+1.4\ [0.0,4.2]\) \\
Qwen3-8B & 512 & Source & 0 & 72 & 72 & 38.5 & \(+0.0\ [-12.2,12.2]\) \\
Qwen3-8B & 1024 & Generic & 1 & 72 & 69 & 67.7 & \(+0.0\ [-12.2,12.2]\) \\
Qwen3-8B & 1024 & Source & 0 & 72 & 72 & 38.5 & \(+0.0\ [-12.2,12.2]\) \\
\bottomrule
\end{tabular}
\end{table*}

%% file: answer-edit-methods.tex
\subsection{Editing a proposed answer on public arithmetic tasks}
\label{sec:answer-edit-methods}

We test whether a model can use its own earlier calculation after its
conclusion has become misleading. We sample 192 questions each from
GSM8K \cite{gsm} and SVAMP \cite{svamp}. The sampling frame contains
distinct prompts with integer targets. The GSM8K sample excludes all
256 items used in the archived arithmetic experiments. The same questions
are shown to Qwen3-4B-Instruct-2507, Qwen3-8B in non-thinking mode,
Gemma 4 E2B, and Gemma 4 E4B.

Each model first generates up to 256 tokens of calculation and then
gives a separate numerical answer. The intervention cohort contains
questions where this initial readout is correct and where a token-length
match is possible. For those questions, we preserve the original
calculation and change one digit of an appended \emph{Proposed answer}.
The replacement keeps the sign and decimal width, avoids leading
zeros, and has the same full review-input token length under that
model's tokenizer. The chosen digit is sampled using a fixed
item-specific order and the token-length constraint. The retained
calculation may mention the original answer. We measure the behavior
of a reviewer given this observable state without assuming that
the entire recorded derivation is sound.

The experiment crosses a correct or edited proposed answer with
the presence or absence of the earlier calculation. We also test
showing only the earlier calculation and a restart with the
question alone. All arms share the same review instruction.
The four answer-containing arms use the same message roles;
the restart drops the earlier assistant message. A separate
numeric instruction reads the result after 64, 192, or 512
tokens from a single independent review trace per condition.
The answer stage has its own 32-token cap. We report actual
review and answer tokens, answer validity, and the frequency
of repeating the inserted number.

The primary contrast is the full-draft penalty of the one-digit
edit at 192 tokens. We report paired exact tests over the eight
model--benchmark cells with Holm adjustment and bootstrap
question-level intervals. When no pair is discordant in a cell, the
absolute accuracy difference is at most the population discordance
rate. Its exact one-sided 95\% upper bound is \(1-.05^{1/n}\) for
\(n\) pairs. We use this bound to give a conservative two-sided
interval in those cells.
For the primary contrast we adjust simultaneous intervals across
those eight cells. The other caps and the factorial interaction
show how the answer penalty varies with budget and the presence
of the earlier work. The same-item intersection across checkpoints
checks sensitivity to model-specific eligibility. A second parser
requires a standalone or boxed result. The intervention estimates
effects on questions the checkpoint initially solved, rather than
accuracy over an unconstrained random benchmark query.

We repeat the unmasked protocol on the 295 distinct integer-answer,
text-only problems in MATH-500 \cite{verify,math500} for Qwen3-8B and
Gemma 4 E4B. This tests answer edits beyond elementary arithmetic.
We report its paired effects separately from the primary
GSM8K/SVAMP family of comparisons.

\paragraph{Removing the proposed answer at readout.}
An answer penalty can propagate into the rewritten calculation or
arise because the old answer remains visible when the model commits.
We use a disjoint sample of 144 questions per benchmark on Qwen3-4B
and Gemma 4 E4B. For each clean or edited full draft, we generate the
review once and read each frozen prefix twice. In one readout the
old proposed answer remains visible. In the other, only that line
is removed. A prefix hash verifies that the new calculation stays
identical within the pair. Comparing the two masked branches
measures any residual dependence on the previously visible answer
when the final readout receives the same original work and sees the
new calculation alone. Comparing with the calculation-only arm
also tests removal before review. We interpret this as a staged
intervention on visible text.

For the masked replay, let \(\Delta_V\) and \(\Delta_M\) denote the
clean-minus-edited accuracy penalties with and without the old
answer visible at readout. Their difference measures the change in
the answer penalty attributable to keeping that line visible for
the final response, with each review prefix unchanged within its
readout pair. \(\Delta_M\) estimates the remaining effect after
the line has been removed. It can be positive if the new review
carried forward information from the misleading answer.

\paragraph{Naturally incorrect answers.}
We separately replay restart, calculation-only context, and the
complete draft on previously screened GSM8K and SVAMP questions
whose initial answer is a parseable wrong integer. Here the old
number is the model's actual answer. We do not insert a correct
answer into those examples. This comparison tests deletion on naturally wrong answers.

%% file: answer-edit-results.tex
\subsection{A misleading answer can coexist with useful work}
\label{sec:answer-edit-results}

The factorial study represents earlier work as a pair of operands.
Here we retain the model's freely generated calculation and edit
only its conclusion. We measure the damage from the edited conclusion
and the value of retaining the calculation once that conclusion
is wrong.

For Qwen3-8B on the integer MATH-500 subset, the 64-token review
answers 84.1\% correctly when given the original calculation with
an edited answer, versus 17.1\% when given the same edited answer
alone and 4.9\% when restarting. Yet relative to the same
calculation with a correct answer, the edit reduces accuracy by
12.2 points (95\% paired CI [$6.1$, $19.5$]). A draft can therefore
be damaged by one answer edit and still be much more valuable than
a clean start at the same incremental budget on these initially solved
problems.

\begin{figure*}[t]
\centering
\includegraphics[width=.98\textwidth]{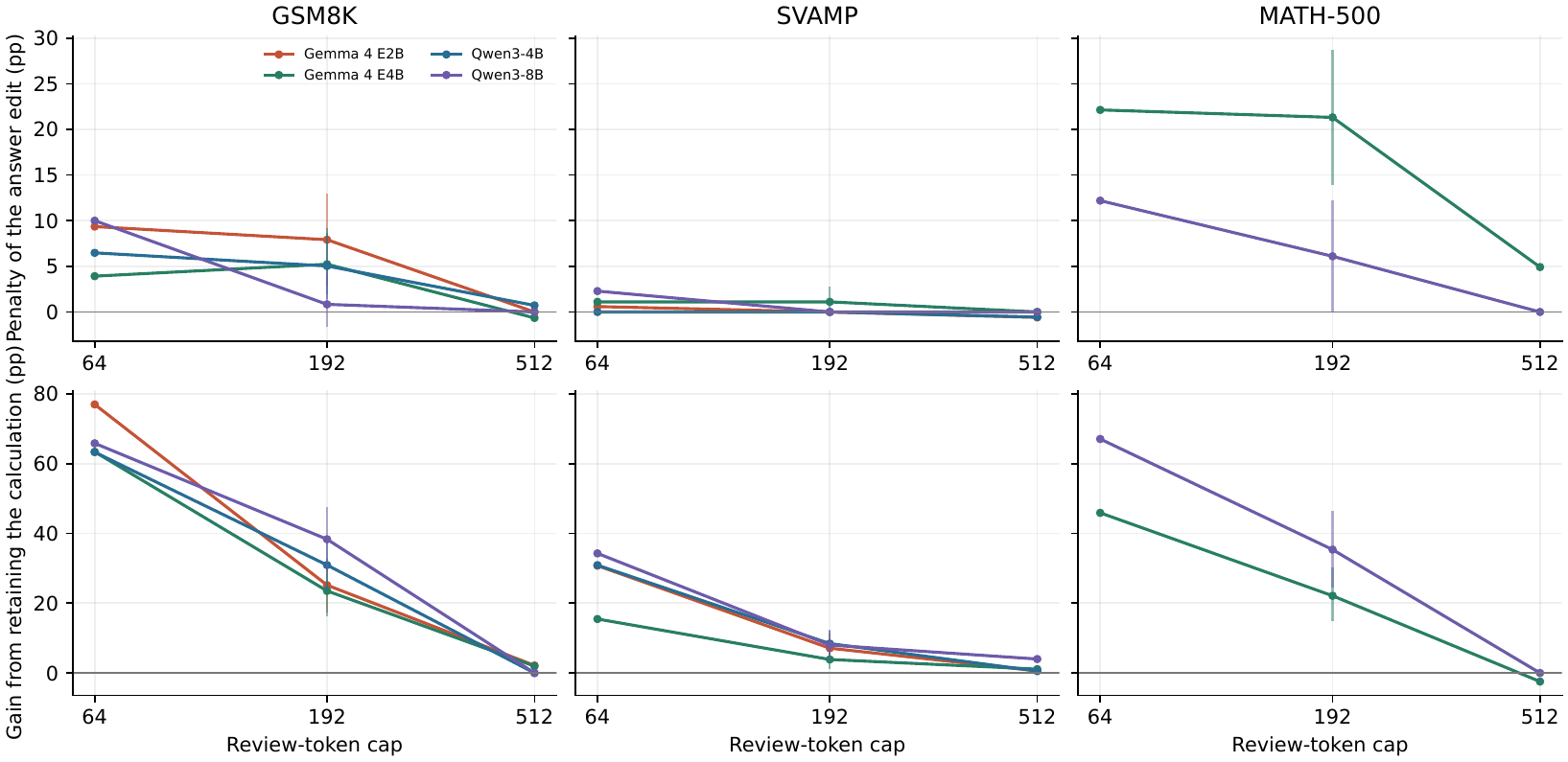}
\caption{Paired effects across review-token caps. The top row compares
correct and edited conclusions with the same calculation. The bottom row
holds the edited conclusion fixed and compares showing or withholding the
earlier calculation. Vertical lines give pointwise 95\% intervals at 192
tokens. Samples are conditioned on an initially correct readout and a
matched edit.}
\label{fig:answer-edit}
\end{figure*}

\begin{table*}[t]
\centering
\small
\caption{One-digit edits at the 192-token reference cap. A penalty is
the clean-minus-edited accuracy gap with the calculation visible. A
retention gain compares showing and withholding the calculation with the
same edited answer. Entries report percentage points and pointwise
paired 95\% intervals. Holm \(p\) values for the penalty use separate
families for the primary benchmarks and for MATH-500. In cells
without discordant pairs, the interval uses an exact binomial bound
on the discordance rate.}
\label{tab:answer-edit}
\begin{tabular}{@{}llrrrr@{}}
\toprule
Benchmark & Checkpoint & Pairs & Edit penalty & Holm \(p\) & Retention gain \\
\midrule
GSM8K & Qwen3-4B & 139 & \(5.0\ [1.4,8.6]\) & .094 & \(30.9\ [23.7,38.8]\) \\
 & Qwen3-8B & 120 & \(0.8\ [-1.7,4.2]\) & 1 & \(38.3\ [29.2,47.5]\) \\
 & Gemma 4 E2B & 139 & \(7.9\ [2.9,12.9]\) & .059 & \(25.2\ [17.3,33.1]\) \\
 & Gemma 4 E4B & 153 & \(5.2\ [2.0,9.2]\) & .059 & \(23.5\ [16.3,30.7]\) \\
\midrule
SVAMP & Qwen3-4B & 178 & \(0.0\ [-1.7,1.7]\) & 1 & \(8.4\ [4.5,12.4]\) \\
 & Qwen3-8B & 175 & \(0.0\ [-1.7,1.7]\) & 1 & \(8.0\ [4.0,12.0]\) \\
 & Gemma 4 E2B & 169 & \(0.0\ [-1.8,1.8]\) & 1 & \(7.1\ [3.0,11.2]\) \\
 & Gemma 4 E4B & 181 & \(1.1\ [0.0,2.8]\) & 1 & \(3.9\ [1.1,6.6]\) \\
\midrule
MATH-500 & Qwen3-8B & 82 & \(6.1\ [0.0,12.2]\) & .125 & \(35.4\ [24.4,46.3]\) \\
 & Gemma 4 E4B & 122 & \(21.3\ [13.9,28.7]\) & \(5.96{\times}10^{-8}\) & \(22.1\ [14.8,30.3]\) \\
\bottomrule
\end{tabular}
\end{table*}

\paragraph{Reusing a draft assumes it already exists.}
On those Qwen3-8B MATH-500 pairs, the 64-cap review with the
calculation and edited answer uses 71.5 generated review and
readout tokens on average. Draft acquisition adds 236.5. A
standalone restart at the 512 cap reaches 98.8\% at 290.4 tokens
on the same selected questions. The short review can make use of
prior work, and its incremental budget omits the cost of obtaining
that work.

\paragraph{The penalty depends on the task and available budget.}
At 192 tokens, all eight primary GSM8K and SVAMP tests for the
full-draft edit have adjusted \(p\ge .059\)
(Table~\ref{tab:answer-edit}). At 512 tokens their observed
penalties range from \(-0.7\) to \(0.7\) points.
Gemma 4 E4B on MATH-500 exhibits a larger 192-token penalty,
21.3 points, which falls to 4.9 at 512. Removing the edited
line before the 192-token review recovers 10.7 points
(95\% paired CI \([4.9,17.2]\)). The strict readout parser gives
the same two 192-token differences.

Holding the wrong number fixed, visible calculation improves all
ten 192-token model--dataset comparisons. On the intersection of
95 GSM8K and 161 SVAMP questions eligible for all four checkpoints,
the retention gains stay positive in all eight comparisons under
the strict parser. Literal exposure to the clean answer is one
reason the calculation may help. In the 59 E4B MATH-500 drafts
without the target as a standalone numeral, keeping the calculation
with the wrong answer still solves 41 cases, compared with 26 when
only the wrong answer appears. This subgroup check leaves open
how much of the gain comes from recoverable computation or
equivalent answer forms.

\paragraph{Visibility at answer time.}
On the disjoint masking sample, the largest 192-token full-draft
penalty is 4.8 points, for E4B on GSM8K. With the same four sets
of review prefixes and the old answer line removed, the clean
and edited branches have no discordant correctness outcomes in
any model--benchmark cell. In the E4B GSM8K cell, masking takes
the edited branch from 95.2\% to 97.1\%, and the clean branch from
100\% to 97.1\%. Thus removing the line erases the observed gap
partly by giving up useful correct information at readout. The
unmasked cell has five discordant pairs (exact \(p=.0625\)).
At 512 tokens both branches solve every eligible item in all
four cells.

\paragraph{Naturally wrong drafts.}
Among 147 model--question pairs with a genuine wrong integer
readout, deleting the answer line before review changes
192-token accuracy by \(-4.3\) to \(+1.4\) points across the four
model--benchmark cells. Each paired interval includes zero.
A naturally wrong first answer can reflect a faulty derivation.
Deletion of the conclusion leaves that derivation in context,
so this study provides a direct boundary for the intervention
on a wrong line appended to initially successful work.

%% file: repair-archived-evidence.tex
\section{Budget dependence in archived repair experiments}
\label{sec:previous-repair-sweeps}

\subsection{Review instructions and budgets}

The archived review sweep uses 36 reserve scenario groups, six per
family, with three count presentations. This design varies the supplied
note and the review instruction, then scores a separate answer readout.
It does not manipulate an independently appended proposed answer. We cross accurate and counterfactual notes with generic and
source review. Each greedy review is scored at 24, 64, and 160 available
tokens through the same numerical answer request. The truncated
prefixes come from the same run and preserve completion if the model
stops early.

A third instruction in the incorrect-note arm requires the two counts
to appear first in a machine-readable line:
\emph{Numerator: N; Denominator: D}. It discourages restatement.
This is a distinct prompt intervention. We check whether the first
matching line supplies the two true counts, and report accuracy
separately for exactly extracted, incorrectly extracted, and unparsed
operands. Outcome differences between those groups are descriptive.
They can reflect problem difficulty as well as which quantities the
model wrote down. All final estimates are still generated by the
model. We exclude groups' individual rows with numerators unchanged by
the complement operation, and resample retained groups for paired
intervals. This follow-up uses the HTML-wrapper-tolerant integer
percentage parser. It reports a difference-in-differences to test
whether the gap between source and generic review changes when the
cap increases from 64 to 160:
\begin{equation}
\begin{split}
I={}&(A_{\mathrm{src},160}-A_{\mathrm{gen},160})\\
    &-(A_{\mathrm{src},64}-A_{\mathrm{gen},64}).
\end{split}
\end{equation}

\begin{figure*}[t]
\centering
\includegraphics[width=.96\textwidth]{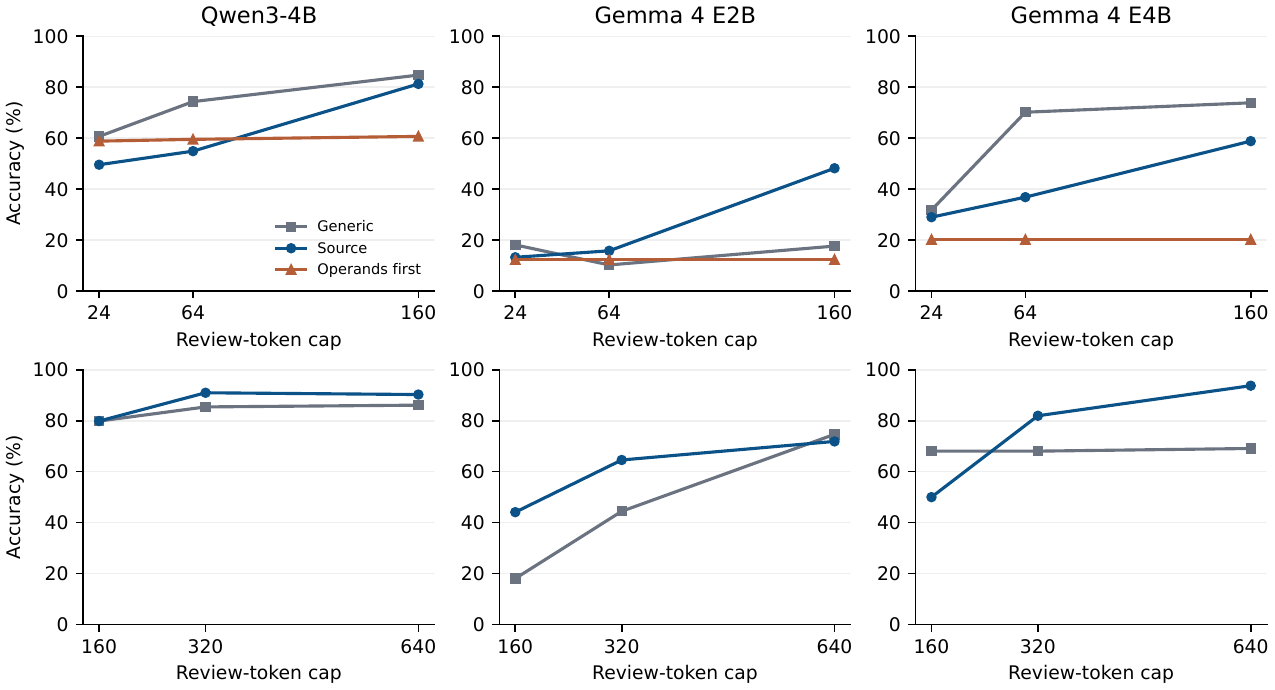}
\caption{Correct answers under corrupted working notes in the archived
studies. The upper row reports the 36-group reserve cohort and the lower
row the separately run 24-group wider-budget subset. Curves join readouts
from one trajectory within each study. The two rows use different
cohorts and generation runs.}
\label{fig:archived-sweep}
\end{figure*}
On the reserve families, Qwen and E4B show a 64-token disadvantage
for source checking ($-19.4$ and $-33.3$ points). Raising the budget to
160 tokens narrows this disadvantage to $-3.5$ and $-15.0$ points,
respectively. E2B's source-check advantage increases from 5.6 to
30.6 points. The paired change in the gap is positive in all three
models (Table~\ref{tab:review-sweep}). This bounds the earlier finding:
an instruction can need more time to realize its benefit, and
increasing the budget can change how alternative instructions rank.

\begin{table}[t]
\centering
\small
\caption{Source minus generic review under corrupted notes, in points.
Interaction is the change in the gap from 64 to 160 available tokens.}
\label{tab:review-sweep}
\begin{tabular}{@{}lrrrr@{}}
\toprule
Model & 24 & 64 & 160 & Interaction [95\% CI] \\
\midrule
Qwen & $-11.1$ & $-19.4$ & $-3.5$ & $16.0$ [$6.7$, $25.0$] \\
E2B & $-4.9$ & $+5.6$ & $+30.6$ & $25.0$ [$11.8$, $38.7$] \\
E4B & $-2.8$ & $-33.3$ & $-15.0$ & $18.3$ [$6.9$, $30.1$] \\
\bottomrule
\end{tabular}
\end{table}

\paragraph{Correct operands can coexist with a wrong answer.}
The operands-first instruction often generates abbreviated labels
\texttt{N:} and \texttt{D:}. The strict literal parser accepts none.
A post-hoc audit of the full recorded 160-token traces also accepts
those abbreviations and simple presentation wrappers. The parser
requires a pair of plain counts and a valid denominator. The model
answer is correct on just 13 of the 84 E2B cases where the emitted
operands are exactly right. E4B answers 38 of 81 such cases correctly,
and Qwen answers 96 of 112. These counts describe the observed
readout in the presence of the earlier corrupt note. They do not
distinguish arithmetic slips from reliance on the earlier note or
other features of the conversation.

\begin{table}[t]
\centering
\small
\caption{Operands-first audit, 206 prompts per model. The last column
reports the paired gain from executing the parsed ratio, with the
original model answer as fallback on unparsed notes.}
\label{tab:operand-audit}
\begin{tabular}{@{}lrrr@{}}
\toprule
Model & Parsed & Exact pair & Execution gain [95\% CI] \\
\midrule
Qwen & 167 & 112 & $3.7$ [$-1.6$, $9.0$] \\
E2B & 205 & 84 & $29.2$ [$15.7$, $42.4$] \\
E4B & 205 & 81 & $19.0$ [$9.3$, $29.6$] \\
\bottomrule
\end{tabular}
\end{table}

Computing $100N/D$ for any parsed pair, with no use of the gold label
by the executor, gives the performance in Table~\ref{tab:operand-audit}.
The comparison includes incorrect extracted operands and falls back to
the stored model answer when it cannot parse a pair. It measures an
offline readout substitution. Its gain does not imply an end-to-end
speedup, and it cannot fix source selection when both extracted
numbers form the wrong ratio.

The same-cap fixed-review comparisons put that gain in context. With exact execution, operands-first reaches
64.4\%, 41.4\%, and 39.1\% for Qwen, E2B, and E4B. The best
observed fixed review for each model reaches 84.7\% (generic),
48.1\% (source), and 73.8\% (generic). The paired execution gaps
against those comparators are $-20.4$ points (95\% CI [$-30.6$, $-10.9$]),
$-6.7$ ([$-22.5$, $8.8$]), and $-34.7$ ([$-50.5$, $-19.0$]).
The gain from substituting the executor diagnoses the final calculation stage.
The full action still depends on the quality of the initial extraction.
\subsection{Review at caps up to 640 tokens}

A separate wider-budget run uses the first 24 of the reserve groups
and retains 140 prompts after excluding identical numerator edits.
It compares the two review instructions on corrupted notes. Each
readout uses a prefix of a single continuation per arm, at 160,
320, or 640 tokens. Accuracy is averaged over scenario groups and
paired uncertainty is calculated with groups as resampling units
(Figure~\ref{fig:archived-sweep}).

On corrupted notes, E4B's source-review effect changes from $-18.1$
points at 160 tokens to $+24.7$ at 640. The paired difference is
$+42.7$ points (95\% CI [$26.7$, $58.7$]). In the opposite direction,
E2B's source advantage drops from $+26.0$ at 160 to $-2.8$ at 640,
because generic review catches up. The 160-token values in
Table~\ref{tab:wide} come from the same subset and longer-run
trajectory as their 640-token comparisons. The action ranking can change as the cap increases on the same notes.

\begin{table}[t]
\centering
\small
\caption{Wider-budget paired differences in accuracy points.
The contrast is source minus generic review.}
\label{tab:wide}
\begin{tabular}{@{}llrrr@{}}
\toprule
Setting & Model & 160 & 320 & 640 \\
\midrule
Review & Qwen & $0.0$ & $+5.6$ & $+4.2$ \\
 & E2B & $+26.0$ & $+20.1$ & $-2.8$ \\
 & E4B & $-18.1$ & $+13.9$ & $+24.7$ \\
\bottomrule
\end{tabular}
\end{table}

The limits also change how much review is realized. In E4B's source
arm, 77.1\% of traces use all 160 tokens, whereas the 640-token cap
yields a mean review length of 275.6 tokens. This establishes that
the cap can interrupt the action, although the aggregate curves
do not locate the exact step at which the review became sufficient.
The archived studies keep the initial note format fixed. The factorial
study adds an answer field and varies its correctness independently.

%% file: repair-discussion.tex
\section{What a successful repair establishes}

\paragraph{Correctness and reuse are different measurements.}
When the proposed answer is wrong, the comparison across operand states
measures the value of having exact counts in the same visible form. When
the operands are wrong, the comparison across answer states measures what
the reviewer can gain from a correct number still present in the draft.
Both interventions concern visible text \cite{turpin,tracefaithfulness}.
The public arithmetic experiment tests a different form of reuse. It
retains the misleading answer while retaining or withholding the larger
piece of written work that accompanied it.

\paragraph{A second pass can repair and regress.}
For a manufactured state with a wrong proposal, any correct output is a
repair relative to Keep. For one with a correct proposal, a wrong or
invalid output is a regression. At 320 tokens the E4B source reviewer
repairs 31 of 48 doubly wrong states and preserves 42 of 48 states with a
wrong pair and a correct proposal. An aggregate across the four arms
weights each of these states equally. The paired outcomes in
Appendix~\ref{app:full-factorial} allow separate comparisons on the
observed strata. A new mixture of naturally occurring states can have
different repair and regression totals.

\paragraph{Commitment is part of a budgeted intervention.}
The staged protocol inserts an answer request after the review prefix.
Single generation requires the reviewer to allocate part of its
continuation to the final answer. The two 31B protocols permit the same
maximum generated-token totals. Their answer instructions and when
those instructions appear differ. The concentration of the 31B
single-generation gap in capped, invalid readouts places the difference
at completion under that protocol. Whether another termination cue
or answer format would produce the same gap remains open.

\paragraph{What the archive adds.}
The natural data reveal an important failure class in the denominator.
One archived note writes \(48/240=20\%\) when \(48/80=60\%\) is the
relevant fraction. There are 240 renewed permits in the entire audit
and 80 permits with the reported property. Running the arithmetic
preserves the wrong conditioning set. The archive identifies such
discrepancies from the source and evaluates the actions offline.
Choosing an action prospectively requires independently observable
diagnostics.

%% file: repair-conclusion.tex
We find no reliable evidence that a weak cue by itself predicts a
greater gain from additional reasoning. Once a draft exists, the
contents of that draft matter. Correct operands can survive a bad
answer, and a correct answer can coexist with faulty operands.
Natural denominator errors motivated direct interventions that change
the denominator, the numerator, or both. At short caps, an action
can reuse intact work quickly and may also carry forward an error.

The multistage experiment shows failures of operand repair on two
checkpoints even when every response in the key comparison is a
valid answer. A source-checking instruction can help one operand
location and hurt another, and the prespecified comparison gives no
reliable overall source advantage. Extending the review to 512 tokens lets 31B recover
all correct proposals in corrupted ledger states. Both Qwen sizes
end Source Review early, typically with the edited operands.
Generic Review on 14B preserves many more correct proposals at
the longer cap. In a different
format, incomplete answers can account for a large accuracy gap.
The same accuracy loss can therefore arise from a failed answer
readout, a repair that needs a longer prefix, or a completed review
that carries a draft error into its answer. Paired state edits,
longer prefixes, and draft-free restarts distinguish these cases
under a fixed checkpoint and question.

%% file: repair-supplement.tex
\onecolumn
\setlength{\LTcapwidth}{.93\textwidth}
\appendix

\section{Reconstruction of the paired experiment}
\label{app:reconstruct}

\subsection{An example of a matched four-state question}

Consider an audit of 400 permits in which 80 have the reported
attribute. Among those 80 there are 48 renewed permits and 32 that were
not renewed. In the other 320 cases there are 192 renewed permits and
128 that were not renewed. Given the attribute, the correct probability
is \(48/80=60\%\). The wrong-pair intervention replaces 48 by 32 so that
the implied probability is \(40\%\). The table shows the four possible
drafts if a 5-point increase to the proposal passes the tokenizer check.
The proposal in the final row is deliberately inconsistent with both
source ratios.

\begin{center}
\begin{tabular}{@{}llrrr@{}}
\toprule
Operands & Proposed answer status & Numerator & Denominator & Proposed answer \\
\midrule
Exact & Correct & 48 & 80 & 60\% \\
Exact & Wrong & 48 & 80 & 65\% \\
Wrong & Correct & 32 & 80 & 60\% \\
Wrong & Wrong & 32 & 80 & 65\% \\
\bottomrule
\end{tabular}
\end{center}

Every state uses the field labels \texttt{Numerator},
\texttt{Denominator}, and \texttt{Proposed answer} on separate lines.
The question stays in the first user message and the intervention
appears in the next user message, after the assistant's draft. The
candidate answer perturbation is chosen in a seeded order from the
admissible \(\pm5\) and \(\pm10\) changes. We compare the realized prompt
lengths under the checkpoint's tokenizer for all four states within
Generic Review and within Source Review. On the confirmation cohort we
also check the two instructions with the appended single-generation cue.
Generating a matched case uses the question and verified operands. It
does not use model performance in the proposed review cells.

\subsection{Action and answer instructions}

\begin{description}
\item[Generic Review] \emph{Review the working note. Check the
calculation and update the answer if needed.}
\item[Source Review] \emph{Review the working note. Re-extract the
matching outcome count and observed count from the original audit, then
calculate.}
\item[Restart] The assistant draft is replaced with \emph{I have the
problem.} The next user turn is \emph{Solve the original question from
the audit.}
\item[Staged answer] The retained continuation is appended after
\texttt{Review:} inside the assistant draft. The user instruction is
\emph{Give the probability as one integer percentage, with its percent
sign. Return only the percentage.}
\item[Single generation] The instruction ends with
\emph{End with a line in the form Final answer: \textless integer\textgreater\%.}
\end{description}

Staged readouts reuse a prefix of the same review at both caps. Single
generation starts a continuation with its own combined token allowance.
The scoring program uses the last line matching the requested final-answer
form for single generation. It scores a missing match as invalid. Keep
and Execute read only the fields of the draft. Execute uses exact
arithmetic and falls back to Keep on an unparseable or inadmissible pair.
Restart runs once per question and cap. Its correctness is shared across
the four factorial states.

\subsection{Units, checks, and reproducibility}

The sampling unit is the scenario group. A group has three alternative
count configurations and two question presentations per configuration.
The screen selects one eligible question per group from the archived E4B
cohort. The confirmation selects one per new group after excluding
question overlap with the archival groups. A single paired bootstrap draw
samples the resulting questions as units. Reusing those indices across
conditions retains correlations among the four draft states, the two
review actions and the two budget limits. The 31B single-generation
comparison uses the same questions as the 31B staged run.
No continuation, draft state, or repeat of the shared restart outcome
is treated as an additional independent item.

The confirmation threshold and estimand were fixed before the screen.
The archived plan specifies a 48-question E4B screen and a single
72-question follow-up conditional on an absolute effect of at least ten
points and at least 85\% sign agreement among paired bootstrap draws.
The staged 320-token contrast is the primary confirmation comparison.
For a constant paired outcome, the reported interval uses an exact
departure-probability bound instead of the degenerate resampling interval.

The artifact stores each question, its draft variants, the review prefix,
the raw response, the parsed estimate, and realized token counts. A
separate archival audit re-extracts \(n,d\) from the rendered source and
checks them against the stored references. Each audit record refers to
a specific archive and row. The supplementary tables are descriptive
tabulations of these records.

\section{Complete factorial results}
\label{app:full-factorial}
Correct, wrong, and invalid outcomes are reported separately. A valid
percentage that differs from the reference counts as wrong. The Full cap
column in the staged condition counts exhaustion of the combined review
and readout allowances. The Source-only and Generic-only counts show
how many paired items drive the accuracy difference.

\input{repair-detail-tables.tex}

\section{Complete natural-draft audit and paired subset}
\label{app:full-natural}
The archive table includes all initial answer states. A ratio can be
numerically correct while its two operands are wrong, so correctness
under Execute and membership in the exact-operand stratum must be
tracked separately. The matched replay assigns a natural draft to an
artificial cell only if both of its initial labels are verifiable.
Examples with an invalid answer or an ambiguous pair remain in the
archive totals.

\input{natural-detail-tables.tex}

\section{Complete arithmetic arm outcomes}
\label{app:full-arithmetic}
The table panels retain the model-specific cohorts, the comparison on
the intersection of eligible questions, MATH-500, the late masking
intervention, and the naturally wrong replays. Eligibility for an
artificial edit requires an initially correct model readout and
matching tokenization. The common cohort holds question membership
fixed across checkpoints within GSM8K and within SVAMP.

The late mask changes which earlier answer is visible at the final
numeric readout while keeping the generated review prefix fixed within
each parent arm. The full-good and full-bad masked arms therefore
keep their respective review trajectories. The work-only arm removes
the proposed answer before the review begins, so that intervention can
also affect how the review unfolds.

\input{public-detail-tables.tex}

\section{Native-thinking budget reference on AIME 2024}
\label{app:native-budget}
The preceding experiments vary visible drafts. This benchmark run
instead follows each model's own native-thinking trace. It provides
a reference for how the relationship between a cap, the realized
continuation, and the final answer changes over a much longer token
range. It contains no controlled operand or answer edits.

\input{repair-native-methods.tex}
\input{native-results.tex}

\subsection{All nine caps and late answer transitions}

Tables~\ref{tab:native-all-counts} and \ref{tab:native-all-usage}
show the full token ladder. Table~\ref{tab:native-all-changes}
also reports whether an item receives new work past the 16,384-token
prefix and whether its answer changes correctness. Length and correctness
are tracked on the same questions.

\input{native-detail-tables.tex}

\section{Two-task outcome and mechanism audit}
\label{app:final-complete}

The additional experiment represents each combination of an operand
edit and an answer-proposal mode in separate rows. An exact state
has correct or wrong proposals. Corrupted states have correct,
coherent wrong or conflicting wrong proposals. Outcome tables count
raw readouts, and the next panels separate repair, regression,
completion and completed-answer accuracy. The trace tables distinguish
a valid new operand pair from reproduction of the specific edit.
The measure captures what the model writes at the start of the review.
It does not establish that a later continuation uses the same operands.

\input{final-two-task-appendix.tex}

\section{Full final ledger budget results}
\label{app:ledger-budget}
The tables report the complete eleven-state design at each budget for
both checkpoints. Keep and Execute are evaluated on the same displayed
states and stay fixed as the generation cap changes. Restart is
evaluated once per question. The archived summary includes realized
length, completion, regression and repair counts for every cell.

\input{ledger-budget-appendix.tex}

\subsection{Qwen3-14B replay with the proposal held correct}
The replay uses the same 24 ledgers, exact state, and three corrupted
operand states with a correct proposal. We reuse the staged non-thinking
protocol and all three caps. Keep preserves all proposals and Execute
is correct only with exact operands. Each action on corrupted operands
has 72 readouts per cap. The descriptive interval resamples questions.
At 1,024 tokens, Source Review accuracy drops by 97.2 points
(95\% CI 93.1--100.0) from exact to corrupted operands.
Among the questions solved by Restart at 1,024, the valid-but-wrong
rate for Source Review is 96.8\% (95\% CI 92.1--100.0).
Across all 72 corrupted states, Source Review begins with the edited
pair in 71 cases and spends a mean of 38.1 review tokens. Generic
Review begins with the edited pair in 70 cases and uses means of
188.3, 498.9 and 797.7 tokens as the cap increases.

\begin{table}[htbp]
\centering\small
\setlength{\tabcolsep}{4pt}
\caption{Qwen3-14B ledger replay. Entries are correct/valid.
Review columns pool the three corrupted states with a correct proposal.
On exact states, both reviews score 24/24 at every cap.}
\label{tab:ledger-qwen14b}
\begin{tabular}{@{}rrrr@{}}
\toprule
Review cap & Generic & Source & Restart \\
\midrule
192   & 32/72 & 2/72 & 0/24 \\
512   & 36/71 & 2/72 & 0/21 \\
1,024 & 63/72 & 2/72 & 21/24 \\
\bottomrule
\end{tabular}
\end{table}

%% file: repair-detail-tables.tex
\subsection{Gemma 4 E4B, staged readout}

\begingroup\small\setlength{\tabcolsep}{3.2pt}
\begin{longtable}{@{}llrrrrrrrr@{}}
\caption{Gemma 4 E4B, staged readout, 64 review tokens. Counts out of 48 per row. Repairs and regressions are paired against Keep. Full cap means total generated tokens equals the sum of the nominal generation allowances. Input range describes the first review prompt. Restart is one shared row per cap.}\label{tab:details-screen-staged-64} \\
\toprule
Operands / answer & Action & Correct & Wrong & Invalid & Repairs & Regress. & Mean gen. & Full cap & Input range \\
\midrule
\endfirsthead
\toprule
Operands / answer & Action & Correct & Wrong & Invalid & Repairs & Regress. & Mean gen. & Full cap & Input range \\
\midrule
\endhead
\midrule
\endfoot
\bottomrule
\endlastfoot
Exact / correct & Keep & 48 & 0 & 0 & 0 & 0 & -- & -- & -- \\
Exact / correct & Execute & 48 & 0 & 0 & 0 & 0 & -- & -- & -- \\
Exact / correct & Generic & 48 & 0 & 0 & 0 & 0 & 68.0 & 0 & 148--168 \\
Exact / correct & Source & 48 & 0 & 0 & 0 & 0 & 68.0 & 0 & 156--176 \\
Exact / wrong & Keep & 0 & 48 & 0 & 0 & 0 & -- & -- & -- \\
Exact / wrong & Execute & 48 & 0 & 0 & 48 & 0 & -- & -- & -- \\
Exact / wrong & Generic & 47 & 1 & 0 & 47 & 0 & 68.0 & 0 & 148--168 \\
Exact / wrong & Source & 48 & 0 & 0 & 48 & 0 & 68.0 & 0 & 156--176 \\
Wrong / correct & Keep & 48 & 0 & 0 & 0 & 0 & -- & -- & -- \\
Wrong / correct & Execute & 0 & 48 & 0 & 0 & 48 & -- & -- & -- \\
Wrong / correct & Generic & 6 & 42 & 0 & 0 & 42 & 68.0 & 0 & 148--168 \\
Wrong / correct & Source & 7 & 40 & 1 & 0 & 41 & 68.0 & 0 & 156--176 \\
Wrong / wrong & Keep & 0 & 48 & 0 & 0 & 0 & -- & -- & -- \\
Wrong / wrong & Execute & 0 & 48 & 0 & 0 & 0 & -- & -- & -- \\
Wrong / wrong & Generic & 5 & 43 & 0 & 5 & 0 & 68.0 & 0 & 148--168 \\
Wrong / wrong & Source & 6 & 41 & 1 & 6 & 0 & 68.0 & 0 & 156--176 \\
Shared & Restart & 7 & 18 & 23 & -- & -- & 69.3 & 0 & 127--146 \\
\end{longtable}\endgroup

\begingroup\small\setlength{\tabcolsep}{3.2pt}
\begin{longtable}{@{}lrrrr@{}}
\caption{Paired Source and Generic Review outcomes at 64 review tokens for Gemma 4 E4B (staged readout).}\label{tab:discordance-screen-staged-64} \\
\toprule
Operands / answer & Both correct & Source only & Generic only & Neither \\
\midrule
\endfirsthead
\toprule
Operands / answer & Both correct & Source only & Generic only & Neither \\
\midrule
\endhead
\midrule
\endfoot
\bottomrule
\endlastfoot
Exact / correct & 48 & 0 & 0 & 0 \\
Exact / wrong & 47 & 1 & 0 & 0 \\
Wrong / correct & 6 & 1 & 0 & 41 \\
Wrong / wrong & 5 & 1 & 0 & 42 \\
\end{longtable}\endgroup

\noindent The Source-minus-Generic gap on wrong proposed answers minus the same gap on correct proposed answers is \(+1.0\ [-2.1,5.2]\) percentage points (paired 95\% interval).

\begingroup\small\setlength{\tabcolsep}{3.2pt}
\begin{longtable}{@{}llrrrrrrrr@{}}
\caption{Gemma 4 E4B, staged readout, 320 review tokens. Counts out of 48 per row. Repairs and regressions are paired against Keep. Full cap means total generated tokens equals the sum of the nominal generation allowances. Input range describes the first review prompt. Restart is one shared row per cap.}\label{tab:details-screen-staged-320} \\
\toprule
Operands / answer & Action & Correct & Wrong & Invalid & Repairs & Regress. & Mean gen. & Full cap & Input range \\
\midrule
\endfirsthead
\toprule
Operands / answer & Action & Correct & Wrong & Invalid & Repairs & Regress. & Mean gen. & Full cap & Input range \\
\midrule
\endhead
\midrule
\endfoot
\bottomrule
\endlastfoot
Exact / correct & Keep & 48 & 0 & 0 & 0 & 0 & -- & -- & -- \\
Exact / correct & Execute & 48 & 0 & 0 & 0 & 0 & -- & -- & -- \\
Exact / correct & Generic & 45 & 2 & 1 & 0 & 3 & 324.0 & 0 & 148--168 \\
Exact / correct & Source & 45 & 0 & 3 & 0 & 3 & 324.2 & 0 & 156--176 \\
Exact / wrong & Keep & 0 & 48 & 0 & 0 & 0 & -- & -- & -- \\
Exact / wrong & Execute & 48 & 0 & 0 & 48 & 0 & -- & -- & -- \\
Exact / wrong & Generic & 41 & 6 & 1 & 41 & 0 & 324.0 & 0 & 148--168 \\
Exact / wrong & Source & 44 & 2 & 2 & 44 & 0 & 324.1 & 0 & 156--176 \\
Wrong / correct & Keep & 48 & 0 & 0 & 0 & 0 & -- & -- & -- \\
Wrong / correct & Execute & 0 & 48 & 0 & 0 & 48 & -- & -- & -- \\
Wrong / correct & Generic & 33 & 14 & 1 & 0 & 15 & 324.1 & 0 & 148--168 \\
Wrong / correct & Source & 42 & 6 & 0 & 0 & 6 & 324.0 & 0 & 156--176 \\
Wrong / wrong & Keep & 0 & 48 & 0 & 0 & 0 & -- & -- & -- \\
Wrong / wrong & Execute & 0 & 48 & 0 & 0 & 0 & -- & -- & -- \\
Wrong / wrong & Generic & 24 & 23 & 1 & 24 & 0 & 324.1 & 0 & 148--168 \\
Wrong / wrong & Source & 31 & 16 & 1 & 31 & 0 & 324.1 & 0 & 156--176 \\
Shared & Restart & 20 & 12 & 16 & -- & -- & 324.8 & 0 & 127--146 \\
\end{longtable}\endgroup

\begingroup\small\setlength{\tabcolsep}{3.2pt}
\begin{longtable}{@{}lrrrr@{}}
\caption{Paired Source and Generic Review outcomes at 320 review tokens for Gemma 4 E4B (staged readout).}\label{tab:discordance-screen-staged-320} \\
\toprule
Operands / answer & Both correct & Source only & Generic only & Neither \\
\midrule
\endfirsthead
\toprule
Operands / answer & Both correct & Source only & Generic only & Neither \\
\midrule
\endhead
\midrule
\endfoot
\bottomrule
\endlastfoot
Exact / correct & 43 & 2 & 2 & 1 \\
Exact / wrong & 38 & 6 & 3 & 1 \\
Wrong / correct & 31 & 11 & 2 & 4 \\
Wrong / wrong & 18 & 13 & 6 & 11 \\
\end{longtable}\endgroup

\noindent The Source-minus-Generic gap on wrong proposed answers minus the same gap on correct proposed answers is \(+1.0\ [-8.3,11.5]\) percentage points (paired 95\% interval).

\begingroup\small\setlength{\tabcolsep}{3.2pt}
\begin{longtable}{@{}lrrrrr@{}}
\caption{Detailed contrasts in percentage points, with paired 95\% intervals. Penalties compare correct with wrong, averaging over the other factor. The interaction subtracts the answer penalty on wrong operands from that on exact operands. The final column is the long-minus-short change in the operand penalty.}\label{tab:effects-screen-staged} \\
\toprule
Action & Cap & Operand penalty & Answer penalty & Operands $\times$ answer & Operand penalty $\Delta$ cap \\
\midrule
\endfirsthead
\toprule
Action & Cap & Operand penalty & Answer penalty & Operands $\times$ answer & Operand penalty $\Delta$ cap \\
\midrule
\endhead
\midrule
\endfoot
\bottomrule
\endlastfoot
Generic & 64 & \(+87.5\ [79.2,94.8]\) & \(+2.1\ [-3.1,7.3]\) & \(+0.0\ [-10.4,10.4]\) & -- \\
Generic & 320 & \(+30.2\ [18.8,41.7]\) & \(+13.5\ [6.2,21.9]\) & \(-10.4\ [-25.0,4.2]\) & \(-57.3\ [-69.8,-44.8]\) \\
Source & 64 & \(+86.5\ [78.1,93.8]\) & \(+1.0\ [-4.2,6.2]\) & \(-2.1\ [-12.5,8.3]\) & -- \\
Source & 320 & \(+16.7\ [6.2,28.1]\) & \(+12.5\ [5.2,20.8]\) & \(-20.8\ [-33.3,-10.4]\) & \(-69.8\ [-82.3,-57.3]\) \\
Restart & 64 & \(+0.0\ [-12.4,12.4]\) & \(+0.0\ [-12.4,12.4]\) & \(+0.0\ [-12.4,12.4]\) & -- \\
Restart & 320 & \(+0.0\ [-12.4,12.4]\) & \(+0.0\ [-12.4,12.4]\) & \(+0.0\ [-12.4,12.4]\) & \(+0.0\ [-12.4,12.4]\) \\
\end{longtable}\endgroup

\subsection{Gemma 4 31B, staged readout}

\begingroup\small\setlength{\tabcolsep}{3.2pt}
\begin{longtable}{@{}llrrrrrrrr@{}}
\caption{Gemma 4 31B, staged readout, 64 review tokens. Counts out of 72 per row. Repairs and regressions are paired against Keep. Full cap means total generated tokens equals the sum of the nominal generation allowances. Input range describes the first review prompt. Restart is one shared row per cap.}\label{tab:details-confirm-staged-64} \\
\toprule
Operands / answer & Action & Correct & Wrong & Invalid & Repairs & Regress. & Mean gen. & Full cap & Input range \\
\midrule
\endfirsthead
\toprule
Operands / answer & Action & Correct & Wrong & Invalid & Repairs & Regress. & Mean gen. & Full cap & Input range \\
\midrule
\endhead
\midrule
\endfoot
\bottomrule
\endlastfoot
Exact / correct & Keep & 72 & 0 & 0 & 0 & 0 & -- & -- & -- \\
Exact / correct & Execute & 72 & 0 & 0 & 0 & 0 & -- & -- & -- \\
Exact / correct & Generic & 72 & 0 & 0 & 0 & 0 & 68.0 & 0 & 151--168 \\
Exact / correct & Source & 72 & 0 & 0 & 0 & 0 & 64.2 & 0 & 159--176 \\
Exact / wrong & Keep & 0 & 72 & 0 & 0 & 0 & -- & -- & -- \\
Exact / wrong & Execute & 72 & 0 & 0 & 72 & 0 & -- & -- & -- \\
Exact / wrong & Generic & 72 & 0 & 0 & 72 & 0 & 68.0 & 0 & 151--168 \\
Exact / wrong & Source & 72 & 0 & 0 & 72 & 0 & 63.8 & 0 & 159--176 \\
Wrong / correct & Keep & 72 & 0 & 0 & 0 & 0 & -- & -- & -- \\
Wrong / correct & Execute & 0 & 72 & 0 & 0 & 72 & -- & -- & -- \\
Wrong / correct & Generic & 67 & 5 & 0 & 0 & 5 & 68.0 & 0 & 151--168 \\
Wrong / correct & Source & 72 & 0 & 0 & 0 & 0 & 68.0 & 0 & 159--176 \\
Wrong / wrong & Keep & 0 & 72 & 0 & 0 & 0 & -- & -- & -- \\
Wrong / wrong & Execute & 0 & 72 & 0 & 0 & 0 & -- & -- & -- \\
Wrong / wrong & Generic & 69 & 3 & 0 & 69 & 0 & 68.0 & 0 & 151--168 \\
Wrong / wrong & Source & 72 & 0 & 0 & 72 & 0 & 67.8 & 0 & 159--176 \\
Shared & Restart & 68 & 4 & 0 & -- & -- & 68.0 & 0 & 130--146 \\
\end{longtable}\endgroup

\begingroup\small\setlength{\tabcolsep}{3.2pt}
\begin{longtable}{@{}lrrrr@{}}
\caption{Paired Source and Generic Review outcomes at 64 review tokens for Gemma 4 31B (staged readout).}\label{tab:discordance-confirm-staged-64} \\
\toprule
Operands / answer & Both correct & Source only & Generic only & Neither \\
\midrule
\endfirsthead
\toprule
Operands / answer & Both correct & Source only & Generic only & Neither \\
\midrule
\endhead
\midrule
\endfoot
\bottomrule
\endlastfoot
Exact / correct & 72 & 0 & 0 & 0 \\
Exact / wrong & 72 & 0 & 0 & 0 \\
Wrong / correct & 67 & 5 & 0 & 0 \\
Wrong / wrong & 69 & 3 & 0 & 0 \\
\end{longtable}\endgroup

\noindent The Source-minus-Generic gap on wrong proposed answers minus the same gap on correct proposed answers is \(-1.4\ [-3.5,0.0]\) percentage points (paired 95\% interval).

\begingroup\small\setlength{\tabcolsep}{3.2pt}
\begin{longtable}{@{}llrrrrrrrr@{}}
\caption{Gemma 4 31B, staged readout, 320 review tokens. Counts out of 72 per row. Repairs and regressions are paired against Keep. Full cap means total generated tokens equals the sum of the nominal generation allowances. Input range describes the first review prompt. Restart is one shared row per cap.}\label{tab:details-confirm-staged-320} \\
\toprule
Operands / answer & Action & Correct & Wrong & Invalid & Repairs & Regress. & Mean gen. & Full cap & Input range \\
\midrule
\endfirsthead
\toprule
Operands / answer & Action & Correct & Wrong & Invalid & Repairs & Regress. & Mean gen. & Full cap & Input range \\
\midrule
\endhead
\midrule
\endfoot
\bottomrule
\endlastfoot
Exact / correct & Keep & 72 & 0 & 0 & 0 & 0 & -- & -- & -- \\
Exact / correct & Execute & 72 & 0 & 0 & 0 & 0 & -- & -- & -- \\
Exact / correct & Generic & 72 & 0 & 0 & 0 & 0 & 231.1 & 0 & 151--168 \\
Exact / correct & Source & 72 & 0 & 0 & 0 & 0 & 177.5 & 0 & 159--176 \\
Exact / wrong & Keep & 0 & 72 & 0 & 0 & 0 & -- & -- & -- \\
Exact / wrong & Execute & 72 & 0 & 0 & 72 & 0 & -- & -- & -- \\
Exact / wrong & Generic & 72 & 0 & 0 & 72 & 0 & 252.1 & 0 & 151--168 \\
Exact / wrong & Source & 72 & 0 & 0 & 72 & 0 & 173.6 & 0 & 159--176 \\
Wrong / correct & Keep & 72 & 0 & 0 & 0 & 0 & -- & -- & -- \\
Wrong / correct & Execute & 0 & 72 & 0 & 0 & 72 & -- & -- & -- \\
Wrong / correct & Generic & 72 & 0 & 0 & 0 & 0 & 302.7 & 0 & 151--168 \\
Wrong / correct & Source & 72 & 0 & 0 & 0 & 0 & 207.2 & 0 & 159--176 \\
Wrong / wrong & Keep & 0 & 72 & 0 & 0 & 0 & -- & -- & -- \\
Wrong / wrong & Execute & 0 & 72 & 0 & 0 & 0 & -- & -- & -- \\
Wrong / wrong & Generic & 72 & 0 & 0 & 72 & 0 & 286.0 & 0 & 151--168 \\
Wrong / wrong & Source & 72 & 0 & 0 & 72 & 0 & 205.8 & 0 & 159--176 \\
Shared & Restart & 72 & 0 & 0 & -- & -- & 294.1 & 0 & 130--146 \\
\end{longtable}\endgroup

\begingroup\small\setlength{\tabcolsep}{3.2pt}
\begin{longtable}{@{}lrrrr@{}}
\caption{Paired Source and Generic Review outcomes at 320 review tokens for Gemma 4 31B (staged readout).}\label{tab:discordance-confirm-staged-320} \\
\toprule
Operands / answer & Both correct & Source only & Generic only & Neither \\
\midrule
\endfirsthead
\toprule
Operands / answer & Both correct & Source only & Generic only & Neither \\
\midrule
\endhead
\midrule
\endfoot
\bottomrule
\endlastfoot
Exact / correct & 72 & 0 & 0 & 0 \\
Exact / wrong & 72 & 0 & 0 & 0 \\
Wrong / correct & 72 & 0 & 0 & 0 \\
Wrong / wrong & 72 & 0 & 0 & 0 \\
\end{longtable}\endgroup

\noindent The Source-minus-Generic gap on wrong proposed answers minus the same gap on correct proposed answers is \(+0.0\ [-8.3,8.3]\) percentage points (paired 95\% interval).

\begingroup\small\setlength{\tabcolsep}{3.2pt}
\begin{longtable}{@{}lrrrrr@{}}
\caption{Detailed contrasts in percentage points, with paired 95\% intervals. Penalties compare correct with wrong, averaging over the other factor. The interaction subtracts the answer penalty on wrong operands from that on exact operands. The final column is the long-minus-short change in the operand penalty.}\label{tab:effects-confirm-staged} \\
\toprule
Action & Cap & Operand penalty & Answer penalty & Operands $\times$ answer & Operand penalty $\Delta$ cap \\
\midrule
\endfirsthead
\toprule
Action & Cap & Operand penalty & Answer penalty & Operands $\times$ answer & Operand penalty $\Delta$ cap \\
\midrule
\endhead
\midrule
\endfoot
\bottomrule
\endlastfoot
Generic & 64 & \(+5.6\ [1.4,11.1]\) & \(-1.4\ [-3.5,0.0]\) & \(+2.8\ [0.0,6.9]\) & -- \\
Generic & 320 & \(+0.0\ [-8.3,8.3]\) & \(+0.0\ [-8.3,8.3]\) & \(+0.0\ [-8.3,8.3]\) & \(-5.6\ [-11.1,-1.4]\) \\
Source & 64 & \(+0.0\ [-8.3,8.3]\) & \(+0.0\ [-8.3,8.3]\) & \(+0.0\ [-8.3,8.3]\) & -- \\
Source & 320 & \(+0.0\ [-8.3,8.3]\) & \(+0.0\ [-8.3,8.3]\) & \(+0.0\ [-8.3,8.3]\) & \(+0.0\ [-8.3,8.3]\) \\
Restart & 64 & \(+0.0\ [-8.3,8.3]\) & \(+0.0\ [-8.3,8.3]\) & \(+0.0\ [-8.3,8.3]\) & -- \\
Restart & 320 & \(+0.0\ [-8.3,8.3]\) & \(+0.0\ [-8.3,8.3]\) & \(+0.0\ [-8.3,8.3]\) & \(+0.0\ [-8.3,8.3]\) \\
\end{longtable}\endgroup

\subsection{Gemma 4 31B, single generation}

\begingroup\small\setlength{\tabcolsep}{3.2pt}
\begin{longtable}{@{}llrrrrrrrr@{}}
\caption{Gemma 4 31B, single generation, 80 total tokens. Counts out of 72 per row. Repairs and regressions are paired against Keep. Full cap means total generated tokens equals the sum of the nominal generation allowances. Input range describes the first review prompt. Restart is one shared row per cap.}\label{tab:details-confirm-normal-64} \\
\toprule
Operands / answer & Action & Correct & Wrong & Invalid & Repairs & Regress. & Mean gen. & Full cap & Input range \\
\midrule
\endfirsthead
\toprule
Operands / answer & Action & Correct & Wrong & Invalid & Repairs & Regress. & Mean gen. & Full cap & Input range \\
\midrule
\endhead
\midrule
\endfoot
\bottomrule
\endlastfoot
Exact / correct & Keep & 72 & 0 & 0 & 0 & 0 & -- & -- & -- \\
Exact / correct & Execute & 72 & 0 & 0 & 0 & 0 & -- & -- & -- \\
Exact / correct & Generic & 0 & 0 & 72 & 0 & 72 & 80.0 & 72 & 166--183 \\
Exact / correct & Source & 0 & 0 & 72 & 0 & 72 & 80.0 & 72 & 174--191 \\
Exact / wrong & Keep & 0 & 72 & 0 & 0 & 0 & -- & -- & -- \\
Exact / wrong & Execute & 72 & 0 & 0 & 72 & 0 & -- & -- & -- \\
Exact / wrong & Generic & 0 & 0 & 72 & 0 & 0 & 80.0 & 72 & 166--183 \\
Exact / wrong & Source & 0 & 0 & 72 & 0 & 0 & 80.0 & 72 & 174--191 \\
Wrong / correct & Keep & 72 & 0 & 0 & 0 & 0 & -- & -- & -- \\
Wrong / correct & Execute & 0 & 72 & 0 & 0 & 72 & -- & -- & -- \\
Wrong / correct & Generic & 0 & 0 & 72 & 0 & 72 & 80.0 & 72 & 166--183 \\
Wrong / correct & Source & 0 & 0 & 72 & 0 & 72 & 80.0 & 72 & 174--191 \\
Wrong / wrong & Keep & 0 & 72 & 0 & 0 & 0 & -- & -- & -- \\
Wrong / wrong & Execute & 0 & 72 & 0 & 0 & 0 & -- & -- & -- \\
Wrong / wrong & Generic & 0 & 0 & 72 & 0 & 0 & 80.0 & 72 & 166--183 \\
Wrong / wrong & Source & 0 & 0 & 72 & 0 & 0 & 80.0 & 72 & 174--191 \\
Shared & Restart & 0 & 0 & 72 & -- & -- & 80.0 & 72 & 145--161 \\
\end{longtable}\endgroup

\begingroup\small\setlength{\tabcolsep}{3.2pt}
\begin{longtable}{@{}lrrrr@{}}
\caption{Paired Source and Generic Review outcomes at 80 total tokens for Gemma 4 31B (single generation).}\label{tab:discordance-confirm-normal-64} \\
\toprule
Operands / answer & Both correct & Source only & Generic only & Neither \\
\midrule
\endfirsthead
\toprule
Operands / answer & Both correct & Source only & Generic only & Neither \\
\midrule
\endhead
\midrule
\endfoot
\bottomrule
\endlastfoot
Exact / correct & 0 & 0 & 0 & 72 \\
Exact / wrong & 0 & 0 & 0 & 72 \\
Wrong / correct & 0 & 0 & 0 & 72 \\
Wrong / wrong & 0 & 0 & 0 & 72 \\
\end{longtable}\endgroup

\noindent The Source-minus-Generic gap on wrong proposed answers minus the same gap on correct proposed answers is \(+0.0\ [-8.3,8.3]\) percentage points (paired 95\% interval).

\begingroup\small\setlength{\tabcolsep}{3.2pt}
\begin{longtable}{@{}llrrrrrrrr@{}}
\caption{Gemma 4 31B, single generation, 336 total tokens. Counts out of 72 per row. Repairs and regressions are paired against Keep. Full cap means total generated tokens equals the sum of the nominal generation allowances. Input range describes the first review prompt. Restart is one shared row per cap.}\label{tab:details-confirm-normal-320} \\
\toprule
Operands / answer & Action & Correct & Wrong & Invalid & Repairs & Regress. & Mean gen. & Full cap & Input range \\
\midrule
\endfirsthead
\toprule
Operands / answer & Action & Correct & Wrong & Invalid & Repairs & Regress. & Mean gen. & Full cap & Input range \\
\midrule
\endhead
\midrule
\endfoot
\bottomrule
\endlastfoot
Exact / correct & Keep & 72 & 0 & 0 & 0 & 0 & -- & -- & -- \\
Exact / correct & Execute & 72 & 0 & 0 & 0 & 0 & -- & -- & -- \\
Exact / correct & Generic & 71 & 0 & 1 & 0 & 1 & 253.8 & 1 & 166--183 \\
Exact / correct & Source & 72 & 0 & 0 & 0 & 0 & 221.4 & 0 & 174--191 \\
Exact / wrong & Keep & 0 & 72 & 0 & 0 & 0 & -- & -- & -- \\
Exact / wrong & Execute & 72 & 0 & 0 & 72 & 0 & -- & -- & -- \\
Exact / wrong & Generic & 70 & 0 & 2 & 70 & 0 & 271.0 & 3 & 166--183 \\
Exact / wrong & Source & 72 & 0 & 0 & 72 & 0 & 216.9 & 0 & 174--191 \\
Wrong / correct & Keep & 72 & 0 & 0 & 0 & 0 & -- & -- & -- \\
Wrong / correct & Execute & 0 & 72 & 0 & 0 & 72 & -- & -- & -- \\
Wrong / correct & Generic & 26 & 0 & 46 & 0 & 46 & 326.8 & 47 & 166--183 \\
Wrong / correct & Source & 72 & 0 & 0 & 0 & 0 & 221.9 & 0 & 174--191 \\
Wrong / wrong & Keep & 0 & 72 & 0 & 0 & 0 & -- & -- & -- \\
Wrong / wrong & Execute & 0 & 72 & 0 & 0 & 0 & -- & -- & -- \\
Wrong / wrong & Generic & 50 & 0 & 22 & 50 & 0 & 311.2 & 26 & 166--183 \\
Wrong / wrong & Source & 72 & 0 & 0 & 72 & 0 & 221.8 & 0 & 174--191 \\
Shared & Restart & 65 & 0 & 7 & -- & -- & 290.9 & 7 & 145--161 \\
\end{longtable}\endgroup

\begingroup\small\setlength{\tabcolsep}{3.2pt}
\begin{longtable}{@{}lrrrr@{}}
\caption{Paired Source and Generic Review outcomes at 336 total tokens for Gemma 4 31B (single generation).}\label{tab:discordance-confirm-normal-320} \\
\toprule
Operands / answer & Both correct & Source only & Generic only & Neither \\
\midrule
\endfirsthead
\toprule
Operands / answer & Both correct & Source only & Generic only & Neither \\
\midrule
\endhead
\midrule
\endfoot
\bottomrule
\endlastfoot
Exact / correct & 71 & 1 & 0 & 0 \\
Exact / wrong & 70 & 2 & 0 & 0 \\
Wrong / correct & 26 & 46 & 0 & 0 \\
Wrong / wrong & 50 & 22 & 0 & 0 \\
\end{longtable}\endgroup

\noindent The Source-minus-Generic gap on wrong proposed answers minus the same gap on correct proposed answers is \(-16.0\ [-23.6,-8.3]\) percentage points (paired 95\% interval).

\begingroup\small\setlength{\tabcolsep}{3.2pt}
\begin{longtable}{@{}lrrrrr@{}}
\caption{Detailed contrasts in percentage points, with paired 95\% intervals. Penalties compare correct with wrong, averaging over the other factor. The interaction subtracts the answer penalty on wrong operands from that on exact operands. The final column is the long-minus-short change in the operand penalty.}\label{tab:effects-confirm-normal} \\
\toprule
Action & Cap & Operand penalty & Answer penalty & Operands $\times$ answer & Operand penalty $\Delta$ cap \\
\midrule
\endfirsthead
\toprule
Action & Cap & Operand penalty & Answer penalty & Operands $\times$ answer & Operand penalty $\Delta$ cap \\
\midrule
\endhead
\midrule
\endfoot
\bottomrule
\endlastfoot
Generic & 80 & \(+0.0\ [-8.3,8.3]\) & \(+0.0\ [-8.3,8.3]\) & \(+0.0\ [-8.3,8.3]\) & -- \\
Generic & 336 & \(+45.1\ [36.8,53.5]\) & \(-16.0\ [-23.6,-8.3]\) & \(+34.7\ [20.8,48.6]\) & \(+45.1\ [36.8,53.5]\) \\
Source & 80 & \(+0.0\ [-8.3,8.3]\) & \(+0.0\ [-8.3,8.3]\) & \(+0.0\ [-8.3,8.3]\) & -- \\
Source & 336 & \(+0.0\ [-8.3,8.3]\) & \(+0.0\ [-8.3,8.3]\) & \(+0.0\ [-8.3,8.3]\) & \(+0.0\ [-8.3,8.3]\) \\
Restart & 80 & \(+0.0\ [-8.3,8.3]\) & \(+0.0\ [-8.3,8.3]\) & \(+0.0\ [-8.3,8.3]\) & -- \\
Restart & 336 & \(+0.0\ [-8.3,8.3]\) & \(+0.0\ [-8.3,8.3]\) & \(+0.0\ [-8.3,8.3]\) & \(+0.0\ [-8.3,8.3]\) \\
\end{longtable}\endgroup

%% file: natural-detail-tables.tex
\begingroup\small\setlength{\tabcolsep}{3.2pt}
\begin{longtable}{@{}llrrrrrrr@{}}
\caption{Full natural-draft audit. Each action entry is correct / valid. The archived review prompts were evaluated at 64 and 160 tokens. Execution uses the parsed ratio and falls back to the initial answer when it cannot parse the operands.}\label{tab:archive-full} \\
\toprule
Operands & Answer & $n$ & Keep & Execute & Generic 64 & Source 64 & Generic 160 & Source 160 \\
\midrule
\endfirsthead
\toprule
Operands & Answer & $n$ & Keep & Execute & Generic 64 & Source 64 & Generic 160 & Source 160 \\
\midrule
\endhead
\midrule
\endfoot
\bottomrule
\endlastfoot
Both & correct & 11 & 11 / 11 & 4 / 11 & 7 / 11 & 6 / 10 & 6 / 10 & 5 / 10 \\
Both & invalid & 35 & 0 / 0 & 0 / 35 & 14 / 25 & 0 / 22 & 16 / 27 & 2 / 25 \\
Both & wrong & 90 & 0 / 90 & 2 / 90 & 33 / 76 & 4 / 78 & 34 / 81 & 12 / 73 \\
Denominator & invalid & 61 & 0 / 0 & 0 / 61 & 24 / 45 & 2 / 20 & 33 / 55 & 21 / 37 \\
Denominator & wrong & 136 & 0 / 136 & 0 / 136 & 42 / 109 & 0 / 117 & 60 / 129 & 45 / 123 \\
Exact & correct & 141 & 141 / 141 & 141 / 141 & 136 / 141 & 138 / 141 & 140 / 141 & 141 / 141 \\
Exact & invalid & 3 & 0 / 0 & 3 / 3 & 3 / 3 & 1 / 3 & 3 / 3 & 1 / 2 \\
Exact & wrong & 2 & 0 / 2 & 2 / 2 & 2 / 2 & 2 / 2 & 2 / 2 & 1 / 1 \\
Numerator & correct & 2 & 2 / 2 & 0 / 2 & 2 / 2 & 2 / 2 & 2 / 2 & 2 / 2 \\
Numerator & wrong & 4 & 0 / 4 & 0 / 4 & 4 / 4 & 1 / 4 & 4 / 4 & 2 / 4 \\
Unscorable & invalid & 12 & 0 / 0 & 0 / 0 & 4 / 11 & 1 / 8 & 4 / 11 & 6 / 11 \\
Unscorable & wrong & 7 & 0 / 7 & 0 / 7 & 1 / 3 & 0 / 7 & 1 / 3 & 6 / 7 \\
\end{longtable}\endgroup

\begingroup\small\setlength{\tabcolsep}{3.2pt}
\begin{longtable}{@{}lllrrrr@{}}
\caption{Matched natural and edited E4B states. Outcome columns are correct counts. The contrast is natural minus edited, in percentage points. The table covers every populated matched stratum. Restart is shared within each comparison.}\label{tab:natural-full} \\
\toprule
Cap & Operands / answer & Action & $n$ & Natural & Edited & Difference [95\% interval] \\
\midrule
\endfirsthead
\toprule
Cap & Operands / answer & Action & $n$ & Natural & Edited & Difference [95\% interval] \\
\midrule
\endhead
\midrule
\endfoot
\bottomrule
\endlastfoot
64 & Exact / correct & Keep & 10 & 10 & 10 & \(+0.0\ [-28.3,28.3]\) \\
64 & Exact / correct & Execute & 10 & 10 & 10 & \(+0.0\ [-28.3,28.3]\) \\
64 & Exact / correct & Generic & 10 & 10 & 10 & \(+0.0\ [-28.3,28.3]\) \\
64 & Exact / correct & Source & 10 & 9 & 10 & \(-10.0\ [-30.0,0.0]\) \\
64 & Exact / correct & Restart & 10 & 6 & 6 & \(+0.0\ [-28.3,28.3]\) \\
64 & Wrong / wrong & Keep & 26 & 0 & 0 & \(+0.0\ [-11.3,11.3]\) \\
64 & Wrong / wrong & Execute & 26 & 0 & 0 & \(+0.0\ [-11.3,11.3]\) \\
64 & Wrong / wrong & Generic & 26 & 0 & 2 & \(-7.7\ [-19.2,0.0]\) \\
64 & Wrong / wrong & Source & 26 & 0 & 2 & \(-7.7\ [-19.2,0.0]\) \\
64 & Wrong / wrong & Restart & 26 & 1 & 1 & \(+0.0\ [-11.3,11.3]\) \\
320 & Exact / correct & Keep & 10 & 10 & 10 & \(+0.0\ [-28.3,28.3]\) \\
320 & Exact / correct & Execute & 10 & 10 & 10 & \(+0.0\ [-28.3,28.3]\) \\
320 & Exact / correct & Generic & 10 & 10 & 10 & \(+0.0\ [-28.3,28.3]\) \\
320 & Exact / correct & Source & 10 & 10 & 10 & \(+0.0\ [-28.3,28.3]\) \\
320 & Exact / correct & Restart & 10 & 9 & 9 & \(+0.0\ [-28.3,28.3]\) \\
320 & Wrong / wrong & Keep & 26 & 0 & 0 & \(+0.0\ [-11.3,11.3]\) \\
320 & Wrong / wrong & Execute & 26 & 0 & 0 & \(+0.0\ [-11.3,11.3]\) \\
320 & Wrong / wrong & Generic & 26 & 7 & 14 & \(-26.9\ [-50.0,0.0]\) \\
320 & Wrong / wrong & Source & 26 & 16 & 18 & \(-7.7\ [-30.8,15.4]\) \\
320 & Wrong / wrong & Restart & 26 & 9 & 9 & \(+0.0\ [-11.3,11.3]\) \\
\end{longtable}\endgroup

%% file: public-detail-tables.tex
\subsection{Public arithmetic answer edits}

\begingroup\small\setlength{\tabcolsep}{3.2pt}
\begin{longtable}{@{}llrlrrrrrr@{}}
\caption{Public arithmetic answer edits. Column groups use review caps of 64, 192 and 512 tokens. E2B and E4B are Gemma 4 checkpoints. Qwen denotes Qwen3. C/S/V gives the standard correct, strict correct and valid counts. T/H gives mean generated tokens including the answer and the percentage of traces at the review scratch cap.}\label{tab:public-public} \\
\toprule
Dataset & Checkpoint & $n$ & Visible state & \shortstack{64\\C/S/V} & \shortstack{64\\T/H} & \shortstack{192\\C/S/V} & \shortstack{192\\T/H} & \shortstack{512\\C/S/V} & \shortstack{512\\T/H} \\
\midrule
\endfirsthead
\toprule
Dataset & Checkpoint & $n$ & Visible state & \shortstack{64\\C/S/V} & \shortstack{64\\T/H} & \shortstack{192\\C/S/V} & \shortstack{192\\T/H} & \shortstack{512\\C/S/V} & \shortstack{512\\T/H} \\
\midrule
\endhead
\midrule
\endfoot
\bottomrule
\endlastfoot
GSM8K & E2B & 139 & Restart & 28/27/139 & 68.2/100 & 102/102/139 & 193.4/91 & 137/137/139 & 303.5/6 \\
GSM8K & E2B & 139 & Work only & 126/126/139 & 67.3/100 & 130/130/139 & 195.3/100 & 139/139/139 & 337.2/4 \\
GSM8K & E2B & 139 & Correct only & 114/114/139 & 67.3/100 & 122/122/139 & 193.6/92 & 137/137/139 & 308.7/6 \\
GSM8K & E2B & 139 & Edited only & 13/13/139 & 67.4/100 & 88/88/139 & 194.3/96 & 136/136/139 & 328.6/9 \\
GSM8K & E2B & 139 & Work + correct & 133/133/139 & 67.3/100 & 134/134/139 & 195.3/99 & 139/139/139 & 335.8/6 \\
GSM8K & E2B & 139 & Work + wrong & 120/120/139 & 67.3/100 & 123/123/139 & 195.3/99 & 139/139/139 & 330.1/4 \\
SVAMP & E2B & 169 & Restart & 131/131/169 & 67.0/100 & 165/165/169 & 180.9/50 & 169/169/169 & 213.0/1 \\
SVAMP & E2B & 169 & Work only & 165/165/168 & 67.2/100 & 168/168/169 & 192.8/87 & 169/169/169 & 256.3/1 \\
SVAMP & E2B & 169 & Correct only & 151/151/169 & 66.9/100 & 169/169/169 & 183.6/59 & 169/169/169 & 227.9/1 \\
SVAMP & E2B & 169 & Edited only & 115/115/169 & 66.9/100 & 156/156/169 & 187.1/72 & 168/168/169 & 271.3/8 \\
SVAMP & E2B & 169 & Work + correct & 168/168/169 & 66.9/100 & 168/168/169 & 191.2/81 & 168/168/169 & 243.1/1 \\
SVAMP & E2B & 169 & Work + wrong & 167/167/169 & 66.9/100 & 168/168/169 & 190.6/78 & 169/169/169 & 243.0/1 \\
GSM8K & E4B & 153 & Restart & 45/45/153 & 67.6/100 & 117/117/153 & 192.5/90 & 151/151/153 & 299.1/3 \\
GSM8K & E4B & 153 & Work only & 149/149/153 & 67.3/100 & 147/147/152 & 195.3/99 & 152/152/153 & 358.6/5 \\
GSM8K & E4B & 153 & Correct only & 114/114/153 & 67.5/100 & 134/134/153 & 192.8/90 & 150/150/153 & 293.8/3 \\
GSM8K & E4B & 153 & Edited only & 48/48/153 & 67.6/100 & 108/108/153 & 192.6/90 & 149/149/153 & 303.7/6 \\
GSM8K & E4B & 153 & Work + correct & 151/151/153 & 67.3/100 & 152/152/153 & 195.1/99 & 151/151/153 & 334.7/3 \\
GSM8K & E4B & 153 & Work + wrong & 145/145/153 & 67.2/100 & 144/144/153 & 194.8/98 & 152/152/153 & 332.9/3 \\
SVAMP & E4B & 181 & Restart & 157/157/180 & 67.1/100 & 176/176/179 & 179.3/45 & 179/179/179 & 204.4/1 \\
SVAMP & E4B & 181 & Work only & 181/181/181 & 66.9/100 & 181/181/181 & 191.1/85 & 180/180/181 & 259.7/3 \\
SVAMP & E4B & 181 & Correct only & 168/168/181 & 66.9/100 & 179/179/181 & 178.6/43 & 181/181/181 & 207.4/2 \\
SVAMP & E4B & 181 & Edited only & 151/151/181 & 66.9/100 & 172/172/179 & 180.6/49 & 179/179/181 & 218.5/3 \\
SVAMP & E4B & 181 & Work + correct & 181/181/181 & 66.9/100 & 181/181/181 & 188.9/76 & 181/181/181 & 244.4/2 \\
SVAMP & E4B & 181 & Work + wrong & 179/179/181 & 66.9/100 & 179/179/181 & 188.5/74 & 181/181/181 & 246.9/2 \\
GSM8K & Qwen 4B & 139 & Restart & 14/14/88 & 82.9/100 & 94/93/135 & 195.0/82 & 137/137/139 & 252.8/0 \\
GSM8K & Qwen 4B & 139 & Work only & 130/130/139 & 67.9/100 & 132/132/139 & 195.5/98 & 139/139/139 & 339.6/4 \\
GSM8K & Qwen 4B & 139 & Correct only & 113/113/135 & 68.1/100 & 124/124/139 & 191.2/83 & 138/138/139 & 254.7/1 \\
GSM8K & Qwen 4B & 139 & Edited only & 39/39/126 & 70.4/100 & 87/87/138 & 195.2/92 & 138/138/139 & 300.3/5 \\
GSM8K & Qwen 4B & 139 & Work + correct & 136/136/139 & 67.4/100 & 137/137/139 & 194.9/99 & 139/139/139 & 321.2/1 \\
GSM8K & Qwen 4B & 139 & Work + wrong & 127/127/139 & 67.9/100 & 130/130/139 & 195.6/98 & 138/138/139 & 311.9/2 \\
SVAMP & Qwen 4B & 178 & Restart & 75/75/152 & 76.9/100 & 160/160/176 & 167.3/42 & 176/176/177 & 209.5/3 \\
SVAMP & Qwen 4B & 178 & Work only & 177/177/178 & 67.1/100 & 178/178/178 & 190.2/82 & 178/178/178 & 304.9/9 \\
SVAMP & Qwen 4B & 178 & Correct only & 166/166/174 & 67.6/100 & 172/172/177 & 169.5/47 & 177/177/177 & 214.7/3 \\
SVAMP & Qwen 4B & 178 & Edited only & 123/123/169 & 68.8/100 & 163/163/176 & 181.7/65 & 177/177/178 & 247.8/6 \\
SVAMP & Qwen 4B & 178 & Work + correct & 178/178/178 & 66.9/100 & 178/178/178 & 189.5/80 & 177/177/177 & 293.9/7 \\
SVAMP & Qwen 4B & 178 & Work + wrong & 178/178/178 & 66.9/100 & 178/178/178 & 193.7/93 & 178/178/178 & 299.1/7 \\
GSM8K & Qwen 8B & 120 & Restart & 4/4/23 & 91.0/100 & 62/59/102 & 197.7/78 & 118/118/119 & 241.1/1 \\
GSM8K & Qwen 8B & 120 & Work only & 114/114/119 & 67.7/100 & 116/116/120 & 194.5/95 & 120/120/120 & 262.2/0 \\
GSM8K & Qwen 8B & 120 & Correct only & 44/44/59 & 82.0/100 & 81/79/113 & 192.6/77 & 118/118/119 & 236.3/1 \\
GSM8K & Qwen 8B & 120 & Edited only & 27/27/50 & 84.0/100 & 72/71/108 & 195.5/78 & 120/120/120 & 238.8/1 \\
GSM8K & Qwen 8B & 120 & Work + correct & 118/118/118 & 67.6/100 & 119/119/120 & 194.4/94 & 120/120/120 & 259.7/0 \\
GSM8K & Qwen 8B & 120 & Work + wrong & 106/106/112 & 69.3/100 & 118/118/120 & 194.0/90 & 120/120/120 & 252.4/0 \\
SVAMP & Qwen 8B & 175 & Restart & 51/48/78 & 83.5/100 & 159/158/168 & 161.8/29 & 171/171/174 & 175.5/1 \\
SVAMP & Qwen 8B & 175 & Work only & 175/174/175 & 66.9/100 & 175/175/175 & 183.7/63 & 175/175/175 & 209.9/1 \\
SVAMP & Qwen 8B & 175 & Correct only & 130/128/139 & 73.2/99 & 168/168/170 & 157.2/27 & 170/170/174 & 169.3/0 \\
SVAMP & Qwen 8B & 175 & Edited only & 110/106/132 & 75.0/100 & 161/161/170 & 162.3/31 & 168/168/175 & 173.9/0 \\
SVAMP & Qwen 8B & 175 & Work + correct & 174/173/175 & 67.0/100 & 175/175/175 & 182.1/59 & 175/175/175 & 208.4/0 \\
SVAMP & Qwen 8B & 175 & Work + wrong & 170/169/171 & 67.7/100 & 175/175/175 & 175.9/55 & 175/175/175 & 199.7/1 \\
\end{longtable}\endgroup

\subsection{Same-question arithmetic cohort}

\begingroup\small\setlength{\tabcolsep}{3.2pt}
\begin{longtable}{@{}llrlrrrrrr@{}}
\caption{Same-question arithmetic cohort. Column groups use review caps of 64, 192 and 512 tokens. E2B and E4B are Gemma 4 checkpoints. Qwen denotes Qwen3. C/S/V gives the standard correct, strict correct and valid counts. T/H gives mean generated tokens including the answer and the percentage of traces at the review scratch cap.}\label{tab:public-common} \\
\toprule
Dataset & Checkpoint & $n$ & Visible state & \shortstack{64\\C/S/V} & \shortstack{64\\T/H} & \shortstack{192\\C/S/V} & \shortstack{192\\T/H} & \shortstack{512\\C/S/V} & \shortstack{512\\T/H} \\
\midrule
\endfirsthead
\toprule
Dataset & Checkpoint & $n$ & Visible state & \shortstack{64\\C/S/V} & \shortstack{64\\T/H} & \shortstack{192\\C/S/V} & \shortstack{192\\T/H} & \shortstack{512\\C/S/V} & \shortstack{512\\T/H} \\
\midrule
\endhead
\midrule
\endfoot
\bottomrule
\endlastfoot
GSM8K & Qwen 4B & 95 & Restart & 12/12/63 & 82.8/100 & 83/82/94 & 190.8/74 & 93/93/95 & 227.1/0 \\
GSM8K & Qwen 4B & 95 & Work only & 95/95/95 & 67.1/100 & 95/95/95 & 194.9/97 & 95/95/95 & 323.0/3 \\
GSM8K & Qwen 4B & 95 & Correct only & 81/81/94 & 67.6/100 & 91/91/95 & 188.8/76 & 95/95/95 & 231.5/0 \\
GSM8K & Qwen 4B & 95 & Edited only & 36/36/86 & 70.6/100 & 74/74/95 & 193.4/88 & 94/94/95 & 269.5/2 \\
GSM8K & Qwen 4B & 95 & Work + correct & 95/95/95 & 67.1/100 & 95/95/95 & 194.4/98 & 95/95/95 & 302.1/1 \\
GSM8K & Qwen 4B & 95 & Work + wrong & 95/95/95 & 67.1/100 & 95/95/95 & 194.8/97 & 94/94/95 & 292.3/2 \\
SVAMP & Qwen 4B & 161 & Restart & 74/74/141 & 75.8/100 & 150/150/159 & 164.5/37 & 161/161/161 & 195.3/1 \\
SVAMP & Qwen 4B & 161 & Work only & 160/160/161 & 67.0/100 & 161/161/161 & 189.6/80 & 161/161/161 & 292.3/6 \\
SVAMP & Qwen 4B & 161 & Correct only & 151/151/158 & 67.4/100 & 158/158/161 & 166.6/42 & 161/161/161 & 199.5/1 \\
SVAMP & Qwen 4B & 161 & Edited only & 117/117/154 & 68.6/100 & 152/152/160 & 180.0/61 & 160/160/161 & 231.9/3 \\
SVAMP & Qwen 4B & 161 & Work + correct & 161/161/161 & 66.8/100 & 161/161/161 & 188.9/78 & 161/161/161 & 281.3/5 \\
SVAMP & Qwen 4B & 161 & Work + wrong & 161/161/161 & 66.8/100 & 161/161/161 & 193.4/93 & 161/161/161 & 288.4/4 \\
GSM8K & Qwen 8B & 95 & Restart & 4/4/21 & 90.3/100 & 61/58/87 & 194.6/72 & 95/95/95 & 228.1/0 \\
GSM8K & Qwen 8B & 95 & Work only & 92/92/94 & 67.7/100 & 92/92/95 & 194.3/94 & 95/95/95 & 256.5/0 \\
GSM8K & Qwen 8B & 95 & Correct only & 42/42/55 & 79.6/100 & 72/70/91 & 189.4/71 & 95/95/95 & 222.5/0 \\
GSM8K & Qwen 8B & 95 & Edited only & 27/27/47 & 81.7/100 & 66/65/88 & 192.7/74 & 95/95/95 & 227.4/0 \\
GSM8K & Qwen 8B & 95 & Work + correct & 95/95/95 & 67.2/100 & 94/94/95 & 194.2/93 & 95/95/95 & 251.5/0 \\
GSM8K & Qwen 8B & 95 & Work + wrong & 85/85/90 & 68.9/100 & 95/95/95 & 193.6/87 & 95/95/95 & 243.4/0 \\
SVAMP & Qwen 8B & 161 & Restart & 50/47/77 & 82.5/100 & 154/153/159 & 158.2/25 & 160/160/161 & 166.2/0 \\
SVAMP & Qwen 8B & 161 & Work only & 161/160/161 & 66.9/100 & 161/161/161 & 182.7/60 & 161/161/161 & 204.8/0 \\
SVAMP & Qwen 8B & 161 & Correct only & 128/126/136 & 71.8/99 & 159/159/159 & 154.0/23 & 158/158/161 & 162.9/0 \\
SVAMP & Qwen 8B & 161 & Edited only & 108/104/129 & 73.7/100 & 153/153/158 & 159.8/28 & 156/156/161 & 169.1/0 \\
SVAMP & Qwen 8B & 161 & Work + correct & 160/159/161 & 67.0/100 & 161/161/161 & 181.4/57 & 161/161/161 & 204.1/0 \\
SVAMP & Qwen 8B & 161 & Work + wrong & 159/158/160 & 67.2/100 & 161/161/161 & 174.2/51 & 161/161/161 & 193.8/0 \\
GSM8K & E2B & 95 & Restart & 27/26/95 & 67.9/100 & 83/83/95 & 192.3/87 & 94/94/95 & 272.4/2 \\
GSM8K & E2B & 95 & Work only & 91/91/95 & 67.2/100 & 93/93/95 & 195.2/100 & 95/95/95 & 316.1/1 \\
GSM8K & E2B & 95 & Correct only & 76/76/95 & 67.3/100 & 88/88/95 & 192.8/91 & 94/94/95 & 281.0/2 \\
GSM8K & E2B & 95 & Edited only & 13/13/95 & 67.3/100 & 75/75/95 & 193.8/95 & 93/93/95 & 300.6/5 \\
GSM8K & E2B & 95 & Work + correct & 93/93/95 & 67.2/100 & 93/93/95 & 195.2/99 & 95/95/95 & 313.6/2 \\
GSM8K & E2B & 95 & Work + wrong & 88/88/95 & 67.2/100 & 90/90/95 & 195.2/99 & 95/95/95 & 304.5/1 \\
SVAMP & E2B & 161 & Restart & 129/129/161 & 66.9/100 & 158/158/161 & 180.1/48 & 161/161/161 & 210.2/1 \\
SVAMP & E2B & 161 & Work only & 157/157/160 & 67.2/100 & 160/160/161 & 192.6/86 & 161/161/161 & 254.4/1 \\
SVAMP & E2B & 161 & Correct only & 144/144/161 & 66.8/100 & 161/161/161 & 183.0/57 & 161/161/161 & 224.8/1 \\
SVAMP & E2B & 161 & Edited only & 114/114/161 & 66.8/100 & 150/150/161 & 186.6/71 & 160/160/161 & 264.0/7 \\
SVAMP & E2B & 161 & Work + correct & 160/160/161 & 66.8/100 & 160/160/161 & 191.0/80 & 160/160/161 & 240.8/1 \\
SVAMP & E2B & 161 & Work + wrong & 159/159/161 & 66.8/100 & 160/160/161 & 190.3/77 & 161/161/161 & 239.5/1 \\
GSM8K & E4B & 95 & Restart & 40/40/95 & 67.6/100 & 87/87/95 & 190.7/84 & 94/94/95 & 270.1/1 \\
GSM8K & E4B & 95 & Work only & 95/95/95 & 67.2/100 & 95/95/95 & 195.2/99 & 95/95/95 & 333.7/2 \\
GSM8K & E4B & 95 & Correct only & 71/71/95 & 67.4/100 & 90/90/95 & 191.3/85 & 94/94/95 & 261.7/0 \\
GSM8K & E4B & 95 & Edited only & 41/41/95 & 67.4/100 & 87/87/95 & 191.0/85 & 94/94/95 & 262.9/0 \\
GSM8K & E4B & 95 & Work + correct & 95/95/95 & 67.2/100 & 95/95/95 & 194.9/98 & 95/95/95 & 302.0/1 \\
GSM8K & E4B & 95 & Work + wrong & 94/94/95 & 67.2/100 & 94/94/95 & 194.5/97 & 95/95/95 & 294.7/0 \\
SVAMP & E4B & 161 & Restart & 144/144/161 & 66.8/100 & 160/160/161 & 178.0/42 & 161/161/161 & 196.6/1 \\
SVAMP & E4B & 161 & Work only & 161/161/161 & 66.8/100 & 161/161/161 & 190.7/84 & 161/161/161 & 250.9/2 \\
SVAMP & E4B & 161 & Correct only & 151/151/161 & 66.8/100 & 160/160/161 & 177.4/40 & 161/161/161 & 198.2/1 \\
SVAMP & E4B & 161 & Edited only & 140/140/161 & 66.8/100 & 158/158/160 & 179.6/47 & 160/160/161 & 208.6/1 \\
SVAMP & E4B & 161 & Work + correct & 161/161/161 & 66.8/100 & 161/161/161 & 188.2/75 & 161/161/161 & 234.0/1 \\
SVAMP & E4B & 161 & Work + wrong & 160/160/161 & 66.8/100 & 161/161/161 & 187.7/71 & 161/161/161 & 235.6/1 \\
\end{longtable}\endgroup

\subsection{Answer edits on MATH-500}

\begingroup\small\setlength{\tabcolsep}{3.2pt}
\begin{longtable}{@{}llrlrrrrrr@{}}
\caption{Answer edits on MATH-500. Column groups use review caps of 64, 192 and 512 tokens. E2B and E4B are Gemma 4 checkpoints. Qwen denotes Qwen3. C/S/V gives the standard correct, strict correct and valid counts. T/H gives mean generated tokens including the answer and the percentage of traces at the review scratch cap.}\label{tab:public-math} \\
\toprule
Dataset & Checkpoint & $n$ & Visible state & \shortstack{64\\C/S/V} & \shortstack{64\\T/H} & \shortstack{192\\C/S/V} & \shortstack{192\\T/H} & \shortstack{512\\C/S/V} & \shortstack{512\\T/H} \\
\midrule
\endfirsthead
\toprule
Dataset & Checkpoint & $n$ & Visible state & \shortstack{64\\C/S/V} & \shortstack{64\\T/H} & \shortstack{192\\C/S/V} & \shortstack{192\\T/H} & \shortstack{512\\C/S/V} & \shortstack{512\\T/H} \\
\midrule
\endhead
\midrule
\endfoot
\bottomrule
\endlastfoot
MATH-500 & E4B & 122 & Restart & 53/53/112 & 68.7/100 & 78/78/115 & 196.8/100 & 117/117/121 & 432.4/41 \\
MATH-500 & E4B & 122 & Work only & 109/109/116 & 67.9/100 & 109/109/117 & 196.0/100 & 118/118/121 & 454.1/43 \\
MATH-500 & E4B & 122 & Correct only & 104/104/121 & 66.9/100 & 110/110/122 & 194.9/100 & 122/122/122 & 435.1/40 \\
MATH-500 & E4B & 122 & Edited only & 38/38/119 & 67.2/100 & 69/69/122 & 195.0/100 & 116/116/122 & 452.7/53 \\
MATH-500 & E4B & 122 & Work + correct & 121/121/122 & 66.8/100 & 122/122/122 & 194.8/100 & 119/119/122 & 451.8/44 \\
MATH-500 & E4B & 122 & Work + wrong & 94/94/121 & 67.0/100 & 96/96/121 & 195.0/100 & 113/113/122 & 452.6/44 \\
MATH-500 & Qwen 8B & 82 & Restart & 4/4/13 & 92.2/100 & 36/35/45 & 206.0/88 & 81/81/82 & 290.4/7 \\
MATH-500 & Qwen 8B & 82 & Work only & 72/72/74 & 70.9/100 & 73/73/78 & 197.0/94 & 82/82/82 & 288.2/6 \\
MATH-500 & Qwen 8B & 82 & Correct only & 24/24/27 & 86.3/100 & 59/59/66 & 197.6/82 & 82/82/82 & 276.0/6 \\
MATH-500 & Qwen 8B & 82 & Edited only & 14/12/19 & 89.9/100 & 46/45/59 & 201.9/87 & 82/82/82 & 289.6/6 \\
MATH-500 & Qwen 8B & 82 & Work + correct & 79/79/79 & 68.9/100 & 80/80/80 & 195.3/91 & 82/82/82 & 285.7/5 \\
MATH-500 & Qwen 8B & 82 & Work + wrong & 69/69/72 & 71.5/100 & 75/75/78 & 195.9/95 & 82/82/82 & 289.9/4 \\
\end{longtable}\endgroup

\subsection{Removing the answer at readout}

\begingroup\small\setlength{\tabcolsep}{3.2pt}
\begin{longtable}{@{}llrlrrrrrr@{}}
\caption{Removing the answer at readout. Column groups use review caps of 64, 192 and 512 tokens. E2B and E4B are Gemma 4 checkpoints. Qwen denotes Qwen3. C/S/V gives the standard correct, strict correct and valid counts. T/H gives mean generated tokens including the answer and the percentage of traces at the review scratch cap.}\label{tab:public-mask} \\
\toprule
Dataset & Checkpoint & $n$ & Visible state & \shortstack{64\\C/S/V} & \shortstack{64\\T/H} & \shortstack{192\\C/S/V} & \shortstack{192\\T/H} & \shortstack{512\\C/S/V} & \shortstack{512\\T/H} \\
\midrule
\endfirsthead
\toprule
Dataset & Checkpoint & $n$ & Visible state & \shortstack{64\\C/S/V} & \shortstack{64\\T/H} & \shortstack{192\\C/S/V} & \shortstack{192\\T/H} & \shortstack{512\\C/S/V} & \shortstack{512\\T/H} \\
\midrule
\endhead
\midrule
\endfoot
\bottomrule
\endlastfoot
GSM8K & E4B & 105 & Work only & 103/103/105 & 67.5/100 & 102/102/105 & 195.5/100 & 105/105/105 & 344.7/2 \\
GSM8K & E4B & 105 & Work + correct & 104/104/105 & 67.5/100 & 105/105/105 & 194.7/96 & 105/105/105 & 317.8/2 \\
GSM8K & E4B & 105 & Work + wrong & 99/99/105 & 67.5/100 & 100/100/105 & 194.6/96 & 105/105/105 & 314.5/2 \\
GSM8K & E4B & 105 & Correct, masked & 103/103/105 & 67.5/100 & 102/102/105 & 194.8/96 & 105/105/105 & 317.8/2 \\
GSM8K & E4B & 105 & Edited, masked & 103/103/105 & 67.5/100 & 102/102/105 & 194.7/96 & 105/105/105 & 314.6/2 \\
SVAMP & E4B & 132 & Work only & 132/132/132 & 66.9/100 & 132/132/132 & 190.9/80 & 132/132/132 & 251.5/1 \\
SVAMP & E4B & 132 & Work + correct & 132/132/132 & 66.9/100 & 132/132/132 & 187.5/70 & 132/132/132 & 235.7/0 \\
SVAMP & E4B & 132 & Work + wrong & 132/132/132 & 66.9/100 & 132/132/132 & 189.1/74 & 132/132/132 & 235.7/0 \\
SVAMP & E4B & 132 & Correct, masked & 132/132/132 & 66.9/100 & 132/132/132 & 187.5/70 & 132/132/132 & 235.7/0 \\
SVAMP & E4B & 132 & Edited, masked & 132/132/132 & 66.9/100 & 132/132/132 & 189.1/74 & 132/132/132 & 235.7/0 \\
GSM8K & Qwen 4B & 92 & Work only & 89/89/92 & 67.3/100 & 91/91/92 & 195.0/97 & 92/92/92 & 323.0/0 \\
GSM8K & Qwen 4B & 92 & Work + correct & 92/92/92 & 67.2/100 & 92/92/92 & 194.3/96 & 92/92/92 & 314.4/0 \\
GSM8K & Qwen 4B & 92 & Work + wrong & 88/88/92 & 67.5/100 & 91/91/92 & 194.9/98 & 92/92/92 & 301.3/0 \\
GSM8K & Qwen 4B & 92 & Correct, masked & 89/89/92 & 67.5/100 & 91/91/92 & 194.5/96 & 92/92/92 & 314.5/0 \\
GSM8K & Qwen 4B & 92 & Edited, masked & 89/89/92 & 67.5/100 & 91/91/92 & 194.9/98 & 92/92/92 & 301.3/0 \\
SVAMP & Qwen 4B & 132 & Work only & 131/131/132 & 66.9/100 & 131/131/132 & 191.6/87 & 132/132/132 & 288.5/4 \\
SVAMP & Qwen 4B & 132 & Work + correct & 132/132/132 & 66.9/100 & 132/132/132 & 190.1/83 & 132/132/132 & 277.9/5 \\
SVAMP & Qwen 4B & 132 & Work + wrong & 132/132/132 & 66.9/100 & 132/132/132 & 193.4/92 & 132/132/132 & 282.5/5 \\
SVAMP & Qwen 4B & 132 & Correct, masked & 131/131/132 & 66.9/100 & 132/132/132 & 190.1/83 & 132/132/132 & 277.9/5 \\
SVAMP & Qwen 4B & 132 & Edited, masked & 131/131/132 & 66.9/100 & 132/132/132 & 193.4/92 & 132/132/132 & 282.5/5 \\
\end{longtable}\endgroup

\subsection{Naturally wrong arithmetic drafts}

\begingroup\small\setlength{\tabcolsep}{3.2pt}
\begin{longtable}{@{}llrlrrrrrr@{}}
\caption{Naturally wrong arithmetic drafts. Column groups use review caps of 64, 192 and 512 tokens. E2B and E4B are Gemma 4 checkpoints. Qwen denotes Qwen3. C/S/V gives the standard correct, strict correct and valid counts. T/H gives mean generated tokens including the answer and the percentage of traces at the review scratch cap.}\label{tab:public-natural-public} \\
\toprule
Dataset & Checkpoint & $n$ & Visible state & \shortstack{64\\C/S/V} & \shortstack{64\\T/H} & \shortstack{192\\C/S/V} & \shortstack{192\\T/H} & \shortstack{512\\C/S/V} & \shortstack{512\\T/H} \\
\midrule
\endfirsthead
\toprule
Dataset & Checkpoint & $n$ & Visible state & \shortstack{64\\C/S/V} & \shortstack{64\\T/H} & \shortstack{192\\C/S/V} & \shortstack{192\\T/H} & \shortstack{512\\C/S/V} & \shortstack{512\\T/H} \\
\midrule
\endhead
\midrule
\endfoot
\bottomrule
\endlastfoot
GSM8K & E4B & 71 & Restart & 6/6/71 & 68.1/100 & 3/3/71 & 196.0/100 & 47/47/71 & 446.9/41 \\
GSM8K & E4B & 71 & Work only & 1/1/71 & 67.9/100 & 4/4/71 & 195.9/100 & 47/47/71 & 461.4/46 \\
GSM8K & E4B & 71 & Work + wrong & 1/1/71 & 67.8/100 & 3/3/71 & 195.8/100 & 48/48/70 & 463.0/45 \\
SVAMP & E4B & 19 & Restart & 2/2/19 & 67.4/100 & 3/3/19 & 189.1/84 & 8/8/19 & 376.5/21 \\
SVAMP & E4B & 19 & Work only & 1/1/19 & 67.4/100 & 2/2/19 & 195.4/100 & 9/9/19 & 435.3/26 \\
SVAMP & E4B & 19 & Work + wrong & 1/1/19 & 67.4/100 & 2/2/19 & 192.2/95 & 6/6/19 & 418.2/26 \\
GSM8K & Qwen 4B & 47 & Restart & 1/1/22 & 84.0/100 & 7/7/40 & 205.3/100 & 31/31/45 & 426.4/38 \\
GSM8K & Qwen 4B & 47 & Work only & 3/3/39 & 77.0/100 & 3/3/45 & 200.6/100 & 27/27/46 & 461.2/55 \\
GSM8K & Qwen 4B & 47 & Work + wrong & 2/2/42 & 72.9/100 & 5/5/44 & 199.4/100 & 29/29/45 & 449.9/45 \\
SVAMP & Qwen 4B & 10 & Restart & 1/1/7 & 75.5/100 & 1/1/9 & 205.3/100 & 1/1/10 & 385.6/20 \\
SVAMP & Qwen 4B & 10 & Work only & 0/0/10 & 72.7/100 & 0/0/10 & 198.3/100 & 1/1/10 & 404.4/30 \\
SVAMP & Qwen 4B & 10 & Work + wrong & 0/0/10 & 69.8/100 & 0/0/10 & 195.4/100 & 2/2/10 & 387.9/20 \\
\end{longtable}\endgroup

%% file: repair-native-methods.tex
\subsection{Models and protocol}

We evaluate native thinking on the
30 AIME 2024 problems distributed by HuggingFaceH4
\cite{aime2024}. We use the same nine caps for each checkpoint:
160, 320, 640, 1{,}024, 2{,}048, 4{,}096, 8{,}192, 16{,}384,
and 32{,}768 tokens. Native Qwen3 and Gemma 4 thinking templates
use their own start and stop markers \cite{qwen3,gemma4}.
The continuation is generated once per question, up to the widest
cap or its natural stop. For a checkpoint readout, we truncate to
the available prefix, close the thinking segment, and prefill a
short final-answer cue. Answer generation has a separate 64-token
cap. The budget is a limit on the thinking segment alone, and a
checkpoint is evaluated even if the natural trajectory had already
ended. Pairing prefixes of the same trajectory holds the decoding
path fixed within each question.

We compare sizes within a release family before comparing families.
The Qwen3 series uses 1.7B, 4B, 8B, and 14B. The Gemma 4 series uses
E2B-IT, E4B-IT, and 31B-IT. The E labels designate effective parameter
counts for those checkpoints. We run BF16 inference with a context
window of 34{,}816 tokens. Gemma 4 31B uses tensor parallelism across
two GPUs to accommodate the same context window. The model families use
different architectures and training recipes.
The observed size trends describe the tested checkpoints. The common
cap defines a nominal generation allowance. FLOPs per token,
sequence segmentation, and memory costs can vary across models.
AIME is public and small. Exact scoring accepts a leading integer or
the last boxed integer in the answer segment. Unparseable answers
count as incorrect. A parsing sensitivity check requires a standalone
integer whenever the answer does not contain a boxed integer.
We report solved counts, paired changes, actual
scratch tokens, and cap saturation. The intervals resample questions
and describe variation across these items with checkpoints and
decoding fixed.

For consecutive caps we count items whose retained prefix actually grows
and break the net score difference into wrong-to-right and right-to-wrong
transitions. We check for changes on identical prefixes to detect
readout variation. A retrospective oracle counts an item as solved if
any measured checkpoint answers it correctly. This uses ground truth
and quantifies the room for choosing among the observed answers.
The replay measures the accuracy and scratch use of each checkpoint
under a shared trajectory. The oracle is a retrospective summary of
changes across readouts.

%% file: native-results.tex
\subsection{Budget Responses in Native Reasoning}
\label{sec:native-results}

\begin{figure*}[t]
\centering
\includegraphics[width=0.96\textwidth]{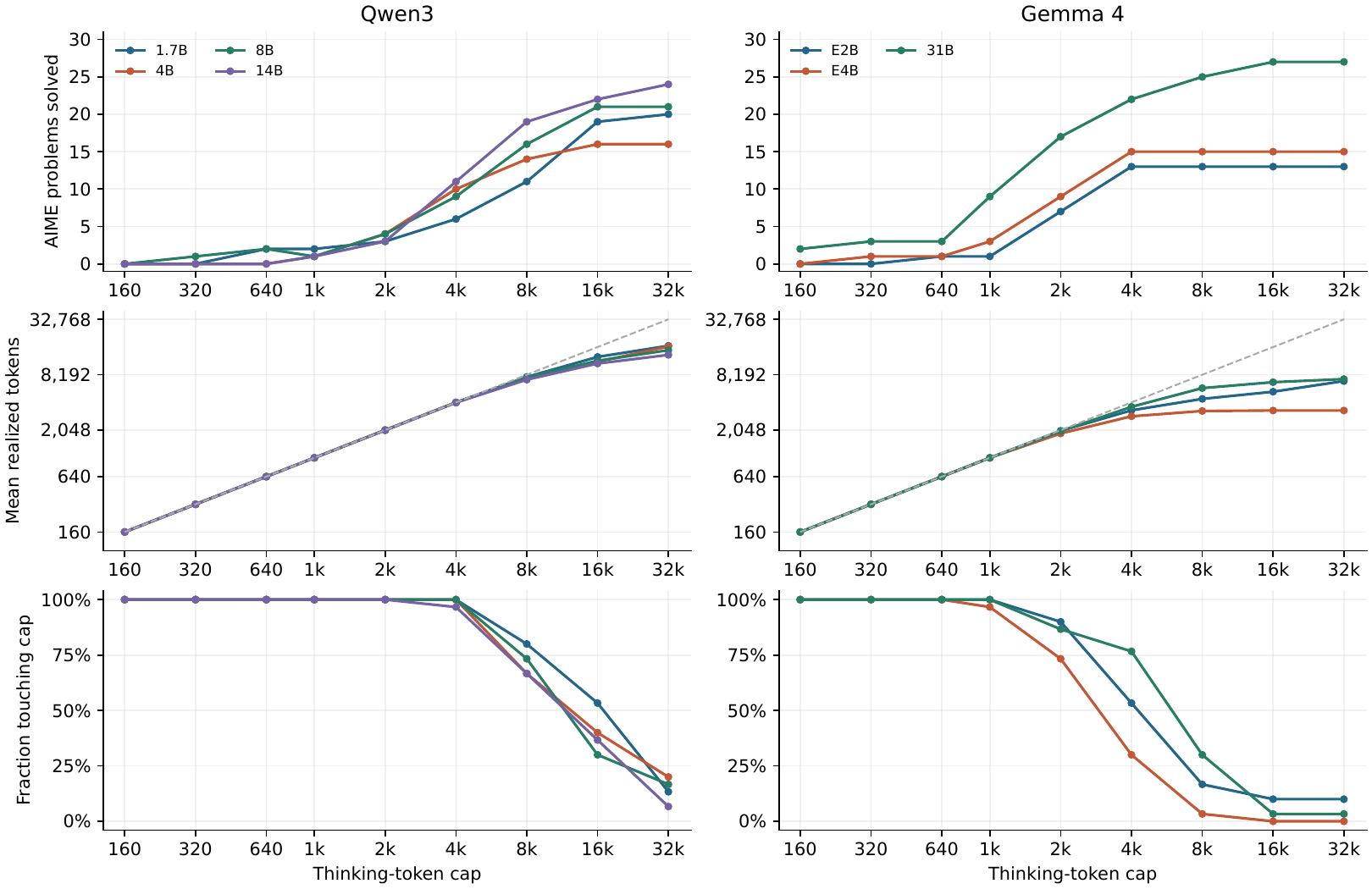}
\caption{Nine thinking-token caps on the same 30 AIME 2024 questions
for each model. The top panels show exact solved counts,
the middle panels show actual scratch use, and the bottom panels
show the fraction of traces that reach the cap. The dotted line is
the available-token ceiling.}
\label{fig:aime-ladder}
\end{figure*}

\begin{table}[t]
\centering
\small
\setlength{\tabcolsep}{3pt}
\caption{Solved AIME 2024 questions out of 30 at selected caps.
Mean scratch and cap hits are measured at 32,768 tokens.}
\label{tab:native-ladder}
\begin{tabular}{@{}lrrrrrrrr@{}}
\toprule
Model & 640 & 2k & 4k & 8k & 16k & 32k & Mean & Hits \\
\midrule
Qwen3 1.7B & 2 & 3 & 6 & 11 & 19 & 20 & 16,905 & 4 \\
Qwen3 4B & 0 & 4 & 10 & 14 & 16 & 16 & 16,601 & 6 \\
Qwen3 8B & 2 & 4 & 9 & 16 & 21 & 21 & 15,136 & 5 \\
Qwen3 14B & 0 & 3 & 11 & 19 & 22 & 24 & 13,462 & 2 \\
\midrule
Gemma 4 E2B & 1 & 7 & 13 & 13 & 13 & 13 & 6,994 & 3 \\
Gemma 4 E4B & 1 & 9 & 15 & 15 & 15 & 15 & 3,346 & 0 \\
Gemma 4 31B & 3 & 17 & 22 & 25 & 27 & 27 & 7,346 & 1 \\
\bottomrule
\end{tabular}
\end{table}

AIME 2024 provides a separate stress test of the budget scale for
native-thinking checkpoints (Figure~\ref{fig:aime-ladder} and
Table~\ref{tab:native-ladder}). Gemma 4 31B already solves 17 of
30 problems at 2{,}048 tokens, compared with 7 for E2B and 9
for E4B. The 31B checkpoint reaches 25 correct at 8{,}192 and
27 at 16{,}384, using a mean of 6{,}800 scratch tokens by the latter
cap. At 32{,}768 the score stays at 27 and only one question extends
past its 16{,}384 prefix. At the shared 8{,}192 cap, 31B solves
6 more items than Qwen3-14B (paired difference of $20.0$ points,
95\% CI [$3.3$, $36.7$]) and uses 5{,}861 versus 7{,}240 scratch
tokens on average. This compares token usage under the tested caps.

Gemma 4 E4B solves 1 of 30 problems at
640 tokens and 15 at 4{,}096. Raising its cap from 8{,}192 to 32{,}768
exposes additional thinking on only one question. Mean realized scratch
increases from 3{,}306 to 3{,}346 tokens, and correctness stays the same
on every question. In this regime the larger allowance primarily
applies to trajectories that have already ended.
The E2B checkpoint also holds its score at 13 from 8{,}192 to
32{,}768, but five questions receive longer traces. Averaged over
those five questions, the increase is 15{,}106 scratch tokens.
Even within Gemma 4, similar plateaus can have very different
costs concentrated in a small tail of questions.

Qwen3-4B illustrates a plateau with additional work. From 16{,}384 to
32{,}768, it generates more scratch on 12 questions and spends
5{,}247 additional tokens per question on average. Under primary
scoring no item changes correctness, and the model solves 16 of 30
at both caps. Requiring a standalone or boxed integer changes
the endpoint counts to 15 and 16. Qwen3-8B also keeps its 16{,}384
score at the widest cap (21 of 30), with 9 prefixes extended and
3{,}512 more tokens per item on average. The realized token
cost distinguishes these outcomes from the E4B plateau.

Qwen3-1.7B has a different response to the same nominal caps. It
solves 6 problems at 4{,}096 tokens and 20 at 32{,}768. Between
8{,}192 and 32{,}768, four wrong integers and five unparseable
readouts turn correct, with no regressions. It generates an
additional 9{,}153 scratch tokens per question on average.
Its accuracy gap relative to E4B
changes from $-30.0$ points at 4{,}096 to $+16.7$ at 32{,}768.
This exploratory cross-checkpoint interaction is $+46.7$ points
(95\% paired CI [$30.0$, $63.3$]). The high-cap score difference
alone has an interval spanning zero. A comparison at one cap
can therefore miss differences in how long the two checkpoints
keep generating and when their additional work pays off.

The 1.7B-versus-4B comparison also crosses within Qwen3. Both use
all 4{,}096 tokens on every item, and 4B leads 10 to 6. At 32{,}768
the smaller checkpoint leads 20 to 16, with comparable mean realized
lengths (16{,}905 and 16{,}601). The cap-by-checkpoint interaction
is $+26.7$ points for 1.7B relative to 4B (95\% paired CI
[$6.7$, $46.7$]). The endpoint interval spans zero. The interaction
describes these two checkpoints and their specific training recipes.
The 14B checkpoint solves the most items of the Qwen series at the
widest cap (24 of 30). It gains five correct answers from 8{,}192
to 32{,}768, while using fewer scratch tokens at that endpoint
than the three smaller Qwen checkpoints.

Individual answers can also get worse at a larger cap. In the same
E4B run, one answer changes from wrong to right and one changes from
right to wrong from 320 to 640 tokens, leaving the aggregate unchanged.
An oracle over all nine readouts solves 16 problems, compared with 15
at the widest cap. This illustrates why equal aggregate accuracy can
conceal either an absence of additional work or offsetting changes.
Across all seven checkpoints, identical prefixes yield the same
parsed answer across all repeated readouts.

%% file: native-detail-tables.tex
\begingroup\small\setlength{\tabcolsep}{3.2pt}
\begin{longtable}{@{}lrrrrrrrrr@{}}
\caption{Correct answers out of 30 at all nine native-thinking caps.}\label{tab:native-all-counts} \\
\toprule
Checkpoint & 160 & 320 & 640 & 1,024 & 2,048 & 4,096 & 8,192 & 16,384 & 32,768 \\
\midrule
\endfirsthead
\toprule
Checkpoint & 160 & 320 & 640 & 1,024 & 2,048 & 4,096 & 8,192 & 16,384 & 32,768 \\
\midrule
\endhead
\midrule
\endfoot
\bottomrule
\endlastfoot
Qwen3 1.7B & 0 & 0 & 2 & 2 & 3 & 6 & 11 & 19 & 20 \\
Qwen3 4B & 0 & 0 & 0 & 1 & 4 & 10 & 14 & 16 & 16 \\
Qwen3 8B & 0 & 1 & 2 & 1 & 4 & 9 & 16 & 21 & 21 \\
Qwen3 14B & 0 & 0 & 0 & 1 & 3 & 11 & 19 & 22 & 24 \\
Gemma 4 E2B & 0 & 0 & 1 & 1 & 7 & 13 & 13 & 13 & 13 \\
Gemma 4 E4B & 0 & 1 & 1 & 3 & 9 & 15 & 15 & 15 & 15 \\
Gemma 4 31B & 2 & 3 & 3 & 9 & 17 & 22 & 25 & 27 & 27 \\
\end{longtable}\endgroup

\begingroup\small\setlength{\tabcolsep}{3.2pt}
\begin{longtable}{@{}lrrrrrrrrr@{}}
\caption{Mean realized native-thinking tokens by checkpoint and cap.}\label{tab:native-all-usage} \\
\toprule
Checkpoint & 160 & 320 & 640 & 1,024 & 2,048 & 4,096 & 8,192 & 16,384 & 32,768 \\
\midrule
\endfirsthead
\toprule
Checkpoint & 160 & 320 & 640 & 1,024 & 2,048 & 4,096 & 8,192 & 16,384 & 32,768 \\
\midrule
\endhead
\midrule
\endfoot
\bottomrule
\endlastfoot
Qwen3 1.7B & 160 & 320 & 640 & 1,024 & 2,048 & 4,096 & 7,752 & 12,798 & 16,905 \\
Qwen3 4B & 160 & 320 & 640 & 1,024 & 2,048 & 4,096 & 7,427 & 11,354 & 16,601 \\
Qwen3 8B & 160 & 320 & 640 & 1,024 & 2,048 & 4,096 & 7,702 & 11,624 & 15,136 \\
Qwen3 14B & 160 & 320 & 640 & 1,024 & 2,048 & 4,079 & 7,240 & 10,839 & 13,462 \\
Gemma 4 E2B & 160 & 320 & 640 & 1,024 & 2,011 & 3,362 & 4,476 & 5,355 & 6,994 \\
Gemma 4 E4B & 160 & 320 & 640 & 1,023 & 1,882 & 2,895 & 3,306 & 3,346 & 3,346 \\
Gemma 4 31B & 160 & 320 & 640 & 1,024 & 1,997 & 3,670 & 5,861 & 6,800 & 7,346 \\
\end{longtable}\endgroup

\begingroup\small\setlength{\tabcolsep}{3.2pt}
\begin{longtable}{@{}lrrrrrr@{}}
\caption{Paired changes from 16,384 to 32,768 tokens. The wrong-to-right and right-to-wrong columns count all correctness flips. The invalid-transition columns identify subsets of those flips. Extra tokens are averaged over all 30 questions.}\label{tab:native-all-changes} \\
\toprule
Checkpoint & Longer prefix & Wrong to right & Right to wrong & Invalid to right & Right to invalid & Extra tokens \\
\midrule
\endfirsthead
\toprule
Checkpoint & Longer prefix & Wrong to right & Right to wrong & Invalid to right & Right to invalid & Extra tokens \\
\midrule
\endhead
\midrule
\endfoot
\bottomrule
\endlastfoot
Qwen3 1.7B & 16 & 2 & 1 & 1 & 0 & 4,107 \\
Qwen3 4B & 12 & 0 & 0 & 0 & 0 & 5,247 \\
Qwen3 8B & 9 & 0 & 0 & 0 & 0 & 3,512 \\
Qwen3 14B & 11 & 2 & 0 & 0 & 0 & 2,623 \\
Gemma 4 E2B & 3 & 0 & 0 & 0 & 0 & 1,638 \\
Gemma 4 E4B & 0 & 0 & 0 & 0 & 0 & 0 \\
Gemma 4 31B & 1 & 0 & 0 & 0 & 0 & 546 \\
\end{longtable}\endgroup

\begingroup\small\setlength{\tabcolsep}{3.2pt}
\begin{longtable}{@{}lrrrrrrrrr@{}}
\caption{Valid numerical readouts out of 30. The gap between these counts and the corresponding counts in Table~\ref{tab:native-all-counts} gives the number of parseable but incorrect answers.}\label{tab:native-all-valid} \\
\toprule
Checkpoint & 160 & 320 & 640 & 1,024 & 2,048 & 4,096 & 8,192 & 16,384 & 32,768 \\
\midrule
\endfirsthead
\toprule
Checkpoint & 160 & 320 & 640 & 1,024 & 2,048 & 4,096 & 8,192 & 16,384 & 32,768 \\
\midrule
\endhead
\midrule
\endfoot
\bottomrule
\endlastfoot
Qwen3 1.7B & 21 & 20 & 20 & 22 & 24 & 23 & 23 & 27 & 27 \\
Qwen3 4B & 27 & 27 & 26 & 24 & 27 & 26 & 27 & 26 & 26 \\
Qwen3 8B & 25 & 23 & 24 & 22 & 22 & 25 & 27 & 29 & 28 \\
Qwen3 14B & 26 & 25 & 25 & 25 & 28 & 28 & 26 & 28 & 29 \\
Gemma 4 E2B & 22 & 25 & 26 & 23 & 27 & 25 & 27 & 28 & 28 \\
Gemma 4 E4B & 26 & 27 & 28 & 26 & 28 & 28 & 28 & 28 & 28 \\
Gemma 4 31B & 24 & 26 & 27 & 28 & 28 & 27 & 29 & 30 & 29 \\
\end{longtable}\endgroup

%% file: final-two-task-appendix.tex
\subsection{Gemma 4 31B, Counts}
\begingroup\small\setlength{\tabcolsep}{3.2pt}
\begin{longtable}{@{}llrrrrrrrr@{}}
\caption{Answer outcomes as correct / completed, out of 24 per cell. K is Keep, E Execute, G Generic Review and S Source Review. Restart counts appear once under each cap.}\label{tab:all-gemma31b-count} \\
\toprule
Operands & Proposal & K64 & E64 & G64 & S64 & K192 & E192 & G192 & S192 \\
\midrule
\endfirsthead
\toprule
Operands & Proposal & K64 & E64 & G64 & S64 & K192 & E192 & G192 & S192 \\
\midrule
\endhead
\midrule
\endfoot
\bottomrule
\endlastfoot
Exact & Correct & 24/24 & 24/24 & 24/24 & 24/24 & 24/24 & 24/24 & 24/24 & 24/24 \\
Exact & Conflict & 0/24 & 24/24 & 24/24 & 24/24 & 0/24 & 24/24 & 24/24 & 24/24 \\
Numerator & Correct & 24/24 & 0/24 & 24/24 & 24/24 & 24/24 & 0/24 & 24/24 & 24/24 \\
Numerator & Conflict & 0/24 & 0/24 & 24/24 & 24/24 & 0/24 & 0/24 & 24/24 & 24/24 \\
Numerator & Coherent & 0/24 & 0/24 & 24/24 & 24/24 & 0/24 & 0/24 & 24/24 & 24/24 \\
Denominator & Correct & 24/24 & 0/24 & 24/24 & 24/24 & 24/24 & 0/24 & 24/24 & 24/24 \\
Denominator & Conflict & 0/24 & 0/24 & 24/24 & 24/24 & 0/24 & 0/24 & 24/24 & 24/24 \\
Denominator & Coherent & 0/24 & 0/24 & 24/24 & 24/24 & 0/24 & 0/24 & 24/24 & 24/24 \\
Both & Correct & 24/24 & 0/24 & 24/24 & 24/24 & 24/24 & 0/24 & 24/24 & 24/24 \\
Both & Conflict & 0/24 & 0/24 & 24/24 & 24/24 & 0/24 & 0/24 & 24/24 & 24/24 \\
Both & Coherent & 0/24 & 0/24 & 24/24 & 24/24 & 0/24 & 0/24 & 24/24 & 24/24 \\
Shared & Restart & 24/24 & -- & -- & -- & 24/24 & -- & -- & -- \\
\end{longtable}\endgroup

\begingroup\small\setlength{\tabcolsep}{3.2pt}
\begin{longtable}{@{}rlrrrrrrr@{}}
\caption{Diagnostic rates as count / denominator. Cells are equally weighted within each initial-answer category. Restart reuses the same results in the two initial-answer categories and is counted once for completion. Review and answer columns show mean realized generated tokens. Input is the mean first prompt length. The last column reports paired effects in points.}\label{tab:rates-gemma31b-count} \\
\toprule
Cap & Action & Repair & Regress & Complete & Correct $\mid$ valid & Review/answer & Input & Coherent minus conflict \\
\midrule
\endfirsthead
\toprule
Cap & Action & Repair & Regress & Complete & Correct $\mid$ valid & Review/answer & Input & Coherent minus conflict \\
\midrule
\endhead
\midrule
\endfoot
\bottomrule
\endlastfoot
64 & Keep & 0/168 & 0/96 & 264/264 & 96/264 & 0.0/0.0 & -- & \(+0.0\ [-12.2,12.2]\) \\
64 & Execute & 24/168 & 72/96 & 264/264 & 48/264 & 0.0/0.0 & -- & \(+0.0\ [-12.2,12.2]\) \\
64 & Generic & 168/168 & 0/96 & 264/264 & 264/264 & 61.5/4.0 & 139.3 & \(+0.0\ [-12.2,12.2]\) \\
64 & Source & 168/168 & 0/96 & 264/264 & 264/264 & 52.7/4.0 & 143.3 & \(+0.0\ [-12.2,12.2]\) \\
64 & Restart & 24/24 & 0/24 & 24/24 & 24/24 & 61.2/4.0 & 118.0 & -- \\
192 & Keep & 0/168 & 0/96 & 264/264 & 96/264 & 0.0/0.0 & -- & \(+0.0\ [-12.2,12.2]\) \\
192 & Execute & 24/168 & 72/96 & 264/264 & 48/264 & 0.0/0.0 & -- & \(+0.0\ [-12.2,12.2]\) \\
192 & Generic & 168/168 & 0/96 & 264/264 & 264/264 & 94.6/4.0 & 139.3 & \(+0.0\ [-12.2,12.2]\) \\
192 & Source & 168/168 & 0/96 & 264/264 & 264/264 & 67.9/4.0 & 143.3 & \(+0.0\ [-12.2,12.2]\) \\
192 & Restart & 24/24 & 0/24 & 24/24 & 24/24 & 97.5/4.0 & 118.0 & -- \\
\end{longtable}\endgroup

\begingroup\small\setlength{\tabcolsep}{3.2pt}
\begin{longtable}{@{}rllrrrr@{}}
\caption{First-line counts among reviews of faulty operands, pooling the three proposal states equally (72 responses per row). The paired question remains the sampling unit.}\label{tab:traces-gemma31b-count} \\
\toprule
Cap & Edited operand & Action & Exact & Repeated edit & Other pair & Unscorable \\
\midrule
\endfirsthead
\toprule
Cap & Edited operand & Action & Exact & Repeated edit & Other pair & Unscorable \\
\midrule
\endhead
\midrule
\endfoot
\bottomrule
\endlastfoot
64 & Numerator & Generic & 72 & 0 & 0 & 0 \\
64 & Numerator & Source & 72 & 0 & 0 & 0 \\
64 & Denominator & Generic & 72 & 0 & 0 & 0 \\
64 & Denominator & Source & 72 & 0 & 0 & 0 \\
64 & Both & Generic & 72 & 0 & 0 & 0 \\
64 & Both & Source & 72 & 0 & 0 & 0 \\
192 & Numerator & Generic & 72 & 0 & 0 & 0 \\
192 & Numerator & Source & 72 & 0 & 0 & 0 \\
192 & Denominator & Generic & 72 & 0 & 0 & 0 \\
192 & Denominator & Source & 72 & 0 & 0 & 0 \\
192 & Both & Generic & 72 & 0 & 0 & 0 \\
192 & Both & Source & 72 & 0 & 0 & 0 \\
\end{longtable}\endgroup

\subsection{Gemma 4 31B, Ledger}
\begingroup\small\setlength{\tabcolsep}{3.2pt}
\begin{longtable}{@{}llrrrrrrrr@{}}
\caption{Answer outcomes as correct / completed, out of 24 per cell. K is Keep, E Execute, G Generic Review and S Source Review. Restart counts appear once under each cap.}\label{tab:all-gemma31b-ledger} \\
\toprule
Operands & Proposal & K64 & E64 & G64 & S64 & K192 & E192 & G192 & S192 \\
\midrule
\endfirsthead
\toprule
Operands & Proposal & K64 & E64 & G64 & S64 & K192 & E192 & G192 & S192 \\
\midrule
\endhead
\midrule
\endfoot
\bottomrule
\endlastfoot
Exact & Correct & 24/24 & 24/24 & 22/24 & 24/24 & 24/24 & 24/24 & 23/24 & 24/24 \\
Exact & Conflict & 0/24 & 24/24 & 22/24 & 21/24 & 0/24 & 24/24 & 21/24 & 20/24 \\
Numerator & Correct & 24/24 & 0/24 & 2/24 & 3/24 & 24/24 & 0/24 & 4/24 & 8/24 \\
Numerator & Conflict & 0/24 & 0/24 & 0/24 & 0/24 & 0/24 & 0/24 & 1/24 & 4/24 \\
Numerator & Coherent & 0/24 & 0/24 & 0/24 & 0/24 & 0/24 & 0/24 & 0/24 & 5/24 \\
Denominator & Correct & 24/24 & 0/24 & 3/24 & 0/24 & 24/24 & 0/24 & 6/24 & 1/24 \\
Denominator & Conflict & 0/24 & 0/24 & 0/24 & 0/24 & 0/24 & 0/24 & 1/24 & 0/24 \\
Denominator & Coherent & 0/24 & 0/24 & 0/24 & 0/24 & 0/24 & 0/24 & 1/24 & 1/24 \\
Both & Correct & 24/24 & 0/24 & 0/24 & 1/24 & 24/24 & 0/24 & 2/24 & 2/24 \\
Both & Conflict & 0/24 & 0/24 & 0/24 & 0/24 & 0/24 & 0/24 & 0/24 & 1/24 \\
Both & Coherent & 0/24 & 0/24 & 0/24 & 0/24 & 0/24 & 0/24 & 0/24 & 1/24 \\
Shared & Restart & 2/24 & -- & -- & -- & 2/24 & -- & -- & -- \\
\end{longtable}\endgroup

\begingroup\small\setlength{\tabcolsep}{3.2pt}
\begin{longtable}{@{}rlrrrrrrr@{}}
\caption{Diagnostic rates as count / denominator. Cells are equally weighted within each initial-answer category. Restart reuses the same results in the two initial-answer categories and is counted once for completion. Review and answer columns show mean realized generated tokens. Input is the mean first prompt length. The last column reports paired effects in points.}\label{tab:rates-gemma31b-ledger} \\
\toprule
Cap & Action & Repair & Regress & Complete & Correct $\mid$ valid & Review/answer & Input & Coherent minus conflict \\
\midrule
\endfirsthead
\toprule
Cap & Action & Repair & Regress & Complete & Correct $\mid$ valid & Review/answer & Input & Coherent minus conflict \\
\midrule
\endhead
\midrule
\endfoot
\bottomrule
\endlastfoot
64 & Keep & 0/168 & 0/96 & 264/264 & 96/264 & 0.0/0.0 & -- & \(+0.0\ [-12.2,12.2]\) \\
64 & Execute & 24/168 & 72/96 & 264/264 & 48/264 & 0.0/0.0 & -- & \(+0.0\ [-12.2,12.2]\) \\
64 & Generic & 22/168 & 69/96 & 264/264 & 49/264 & 64.0/4.0 & 363.5 & \(+0.0\ [-12.2,12.2]\) \\
64 & Source & 21/168 & 68/96 & 264/264 & 49/264 & 64.0/4.0 & 367.5 & \(+0.0\ [-12.2,12.2]\) \\
64 & Restart & 2/24 & 22/24 & 24/24 & 2/24 & 64.0/4.0 & 342.2 & -- \\
192 & Keep & 0/168 & 0/96 & 264/264 & 96/264 & 0.0/0.0 & -- & \(+0.0\ [-12.2,12.2]\) \\
192 & Execute & 24/168 & 72/96 & 264/264 & 48/264 & 0.0/0.0 & -- & \(+0.0\ [-12.2,12.2]\) \\
192 & Generic & 24/168 & 61/96 & 264/264 & 59/264 & 192.0/4.0 & 363.5 & \(-1.4\ [-4.2,0.0]\) \\
192 & Source & 32/168 & 61/96 & 264/264 & 67/264 & 192.0/4.0 & 367.5 & \(+2.8\ [0.0,6.9]\) \\
192 & Restart & 2/24 & 22/24 & 24/24 & 2/24 & 192.0/4.0 & 342.2 & -- \\
\end{longtable}\endgroup

\begingroup\small\setlength{\tabcolsep}{3.2pt}
\begin{longtable}{@{}rllrrrr@{}}
\caption{First-line counts among reviews of faulty operands, pooling the three proposal states equally (72 responses per row). The paired question remains the sampling unit.}\label{tab:traces-gemma31b-ledger} \\
\toprule
Cap & Edited operand & Action & Exact & Repeated edit & Other pair & Unscorable \\
\midrule
\endfirsthead
\toprule
Cap & Edited operand & Action & Exact & Repeated edit & Other pair & Unscorable \\
\midrule
\endhead
\midrule
\endfoot
\bottomrule
\endlastfoot
64 & Numerator & Generic & 1 & 46 & 25 & 0 \\
64 & Numerator & Source & 0 & 49 & 23 & 0 \\
64 & Denominator & Generic & 1 & 20 & 51 & 0 \\
64 & Denominator & Source & 1 & 42 & 29 & 0 \\
64 & Both & Generic & 0 & 26 & 46 & 0 \\
64 & Both & Source & 0 & 39 & 33 & 0 \\
192 & Numerator & Generic & 1 & 46 & 25 & 0 \\
192 & Numerator & Source & 0 & 49 & 23 & 0 \\
192 & Denominator & Generic & 1 & 20 & 51 & 0 \\
192 & Denominator & Source & 1 & 42 & 29 & 0 \\
192 & Both & Generic & 0 & 26 & 46 & 0 \\
192 & Both & Source & 0 & 39 & 33 & 0 \\
\end{longtable}\endgroup

\subsection{Qwen3-8B, Counts}
\begingroup\small\setlength{\tabcolsep}{3.2pt}
\begin{longtable}{@{}llrrrrrrrr@{}}
\caption{Answer outcomes as correct / completed, out of 24 per cell. K is Keep, E Execute, G Generic Review and S Source Review. Restart counts appear once under each cap.}\label{tab:all-qwen8b-count} \\
\toprule
Operands & Proposal & K64 & E64 & G64 & S64 & K192 & E192 & G192 & S192 \\
\midrule
\endfirsthead
\toprule
Operands & Proposal & K64 & E64 & G64 & S64 & K192 & E192 & G192 & S192 \\
\midrule
\endhead
\midrule
\endfoot
\bottomrule
\endlastfoot
Exact & Correct & 24/24 & 24/24 & 23/24 & 24/24 & 24/24 & 24/24 & 23/24 & 24/24 \\
Exact & Conflict & 0/24 & 24/24 & 23/24 & 24/24 & 0/24 & 24/24 & 23/24 & 24/24 \\
Numerator & Correct & 24/24 & 0/24 & 24/24 & 21/24 & 24/24 & 0/24 & 24/24 & 21/24 \\
Numerator & Conflict & 0/24 & 0/24 & 23/24 & 22/24 & 0/24 & 0/24 & 23/24 & 22/24 \\
Numerator & Coherent & 0/24 & 0/24 & 23/24 & 21/24 & 0/24 & 0/24 & 23/24 & 21/24 \\
Denominator & Correct & 24/24 & 0/24 & 12/24 & 14/24 & 24/24 & 0/24 & 12/24 & 14/24 \\
Denominator & Conflict & 0/24 & 0/24 & 12/24 & 12/24 & 0/24 & 0/24 & 12/24 & 12/24 \\
Denominator & Coherent & 0/24 & 0/24 & 11/24 & 10/24 & 0/24 & 0/24 & 11/24 & 10/24 \\
Both & Correct & 24/24 & 0/24 & 22/24 & 21/24 & 24/24 & 0/24 & 22/24 & 21/24 \\
Both & Conflict & 0/24 & 0/24 & 23/24 & 21/24 & 0/24 & 0/24 & 23/24 & 21/24 \\
Both & Coherent & 0/24 & 0/24 & 21/24 & 21/24 & 0/24 & 0/24 & 21/24 & 21/24 \\
Shared & Restart & 23/24 & -- & -- & -- & 23/24 & -- & -- & -- \\
\end{longtable}\endgroup

\begingroup\small\setlength{\tabcolsep}{3.2pt}
\begin{longtable}{@{}rlrrrrrrr@{}}
\caption{Diagnostic rates as count / denominator. Cells are equally weighted within each initial-answer category. Restart reuses the same results in the two initial-answer categories and is counted once for completion. Review and answer columns show mean realized generated tokens. Input is the mean first prompt length. The last column reports paired effects in points.}\label{tab:rates-qwen8b-count} \\
\toprule
Cap & Action & Repair & Regress & Complete & Correct $\mid$ valid & Review/answer & Input & Coherent minus conflict \\
\midrule
\endfirsthead
\toprule
Cap & Action & Repair & Regress & Complete & Correct $\mid$ valid & Review/answer & Input & Coherent minus conflict \\
\midrule
\endhead
\midrule
\endfoot
\bottomrule
\endlastfoot
64 & Keep & 0/168 & 0/96 & 264/264 & 96/264 & 0.0/0.0 & -- & \(+0.0\ [-12.2,12.2]\) \\
64 & Execute & 24/168 & 72/96 & 264/264 & 48/264 & 0.0/0.0 & -- & \(+0.0\ [-12.2,12.2]\) \\
64 & Generic & 136/168 & 15/96 & 264/264 & 217/264 & 50.5/4.0 & 140.3 & \(-4.2\ [-9.7,1.4]\) \\
64 & Source & 131/168 & 16/96 & 264/264 & 211/264 & 44.0/4.0 & 144.3 & \(-4.2\ [-11.1,2.8]\) \\
64 & Restart & 23/24 & 1/24 & 24/24 & 23/24 & 41.0/4.0 & 117.0 & -- \\
192 & Keep & 0/168 & 0/96 & 264/264 & 96/264 & 0.0/0.0 & -- & \(+0.0\ [-12.2,12.2]\) \\
192 & Execute & 24/168 & 72/96 & 264/264 & 48/264 & 0.0/0.0 & -- & \(+0.0\ [-12.2,12.2]\) \\
192 & Generic & 136/168 & 15/96 & 264/264 & 217/264 & 50.8/4.0 & 140.3 & \(-4.2\ [-9.7,1.4]\) \\
192 & Source & 131/168 & 16/96 & 264/264 & 211/264 & 44.0/4.0 & 144.3 & \(-4.2\ [-11.1,2.8]\) \\
192 & Restart & 23/24 & 1/24 & 24/24 & 23/24 & 41.0/4.0 & 117.0 & -- \\
\end{longtable}\endgroup

\begingroup\small\setlength{\tabcolsep}{3.2pt}
\begin{longtable}{@{}rllrrrr@{}}
\caption{First-line counts among reviews of faulty operands, pooling the three proposal states equally (72 responses per row). The paired question remains the sampling unit.}\label{tab:traces-qwen8b-count} \\
\toprule
Cap & Edited operand & Action & Exact & Repeated edit & Other pair & Unscorable \\
\midrule
\endfirsthead
\toprule
Cap & Edited operand & Action & Exact & Repeated edit & Other pair & Unscorable \\
\midrule
\endhead
\midrule
\endfoot
\bottomrule
\endlastfoot
64 & Numerator & Generic & 69 & 1 & 1 & 1 \\
64 & Numerator & Source & 46 & 0 & 3 & 23 \\
64 & Denominator & Generic & 35 & 34 & 3 & 0 \\
64 & Denominator & Source & 25 & 23 & 0 & 24 \\
64 & Both & Generic & 65 & 0 & 6 & 1 \\
64 & Both & Source & 28 & 0 & 9 & 35 \\
192 & Numerator & Generic & 69 & 1 & 1 & 1 \\
192 & Numerator & Source & 46 & 0 & 3 & 23 \\
192 & Denominator & Generic & 35 & 34 & 3 & 0 \\
192 & Denominator & Source & 25 & 23 & 0 & 24 \\
192 & Both & Generic & 65 & 0 & 6 & 1 \\
192 & Both & Source & 28 & 0 & 9 & 35 \\
\end{longtable}\endgroup

\subsection{Qwen3-8B, Ledger}
\begingroup\small\setlength{\tabcolsep}{3.2pt}
\begin{longtable}{@{}llrrrrrrrr@{}}
\caption{Answer outcomes as correct / completed, out of 24 per cell. K is Keep, E Execute, G Generic Review and S Source Review. Restart counts appear once under each cap.}\label{tab:all-qwen8b-ledger} \\
\toprule
Operands & Proposal & K64 & E64 & G64 & S64 & K192 & E192 & G192 & S192 \\
\midrule
\endfirsthead
\toprule
Operands & Proposal & K64 & E64 & G64 & S64 & K192 & E192 & G192 & S192 \\
\midrule
\endhead
\midrule
\endfoot
\bottomrule
\endlastfoot
Exact & Correct & 24/24 & 24/24 & 24/24 & 24/24 & 24/24 & 24/24 & 24/24 & 24/24 \\
Exact & Conflict & 0/24 & 24/24 & 23/24 & 24/24 & 0/24 & 24/24 & 23/24 & 24/24 \\
Numerator & Correct & 24/24 & 0/24 & 0/24 & 0/24 & 24/24 & 0/24 & 0/24 & 0/24 \\
Numerator & Conflict & 0/24 & 0/24 & 0/24 & 0/24 & 0/24 & 0/24 & 0/24 & 0/24 \\
Numerator & Coherent & 0/24 & 0/24 & 0/24 & 0/24 & 0/24 & 0/24 & 0/24 & 0/24 \\
Denominator & Correct & 24/24 & 0/24 & 1/24 & 0/24 & 24/24 & 0/24 & 1/24 & 0/24 \\
Denominator & Conflict & 0/24 & 0/24 & 0/24 & 0/24 & 0/24 & 0/24 & 0/24 & 0/24 \\
Denominator & Coherent & 0/24 & 0/24 & 0/24 & 0/24 & 0/24 & 0/24 & 0/24 & 0/24 \\
Both & Correct & 24/24 & 0/24 & 1/24 & 0/24 & 24/24 & 0/24 & 0/24 & 0/24 \\
Both & Conflict & 0/24 & 0/24 & 0/24 & 0/24 & 0/24 & 0/24 & 0/24 & 0/24 \\
Both & Coherent & 0/24 & 0/24 & 0/24 & 0/24 & 0/24 & 0/24 & 0/24 & 0/24 \\
Shared & Restart & 0/1 & -- & -- & -- & 0/2 & -- & -- & -- \\
\end{longtable}\endgroup

\begingroup\small\setlength{\tabcolsep}{3.2pt}
\begin{longtable}{@{}rlrrrrrrr@{}}
\caption{Diagnostic rates as count / denominator. Cells are equally weighted within each initial-answer category. Restart reuses the same results in the two initial-answer categories and is counted once for completion. Review and answer columns show mean realized generated tokens. Input is the mean first prompt length. The last column reports paired effects in points.}\label{tab:rates-qwen8b-ledger} \\
\toprule
Cap & Action & Repair & Regress & Complete & Correct $\mid$ valid & Review/answer & Input & Coherent minus conflict \\
\midrule
\endfirsthead
\toprule
Cap & Action & Repair & Regress & Complete & Correct $\mid$ valid & Review/answer & Input & Coherent minus conflict \\
\midrule
\endhead
\midrule
\endfoot
\bottomrule
\endlastfoot
64 & Keep & 0/168 & 0/96 & 264/264 & 96/264 & 0.0/0.0 & -- & \(+0.0\ [-12.2,12.2]\) \\
64 & Execute & 24/168 & 72/96 & 264/264 & 48/264 & 0.0/0.0 & -- & \(+0.0\ [-12.2,12.2]\) \\
64 & Generic & 23/168 & 70/96 & 264/264 & 49/264 & 52.1/4.0 & 373.5 & \(+0.0\ [-12.2,12.2]\) \\
64 & Source & 24/168 & 72/96 & 264/264 & 48/264 & 38.6/4.0 & 377.5 & \(+0.0\ [-12.2,12.2]\) \\
64 & Restart & 0/24 & 24/24 & 1/24 & 0/1 & 64.0/15.5 & 350.2 & -- \\
192 & Keep & 0/168 & 0/96 & 264/264 & 96/264 & 0.0/0.0 & -- & \(+0.0\ [-12.2,12.2]\) \\
192 & Execute & 24/168 & 72/96 & 264/264 & 48/264 & 0.0/0.0 & -- & \(+0.0\ [-12.2,12.2]\) \\
192 & Generic & 23/168 & 71/96 & 264/264 & 48/264 & 56.1/4.0 & 373.5 & \(+0.0\ [-12.2,12.2]\) \\
192 & Source & 24/168 & 72/96 & 264/264 & 48/264 & 38.6/4.0 & 377.5 & \(+0.0\ [-12.2,12.2]\) \\
192 & Restart & 0/24 & 24/24 & 2/24 & 0/2 & 192.0/15.0 & 350.2 & -- \\
\end{longtable}\endgroup

\begingroup\small\setlength{\tabcolsep}{3.2pt}
\begin{longtable}{@{}rllrrrr@{}}
\caption{First-line counts among reviews of faulty operands, pooling the three proposal states equally (72 responses per row). The paired question remains the sampling unit.}\label{tab:traces-qwen8b-ledger} \\
\toprule
Cap & Edited operand & Action & Exact & Repeated edit & Other pair & Unscorable \\
\midrule
\endfirsthead
\toprule
Cap & Edited operand & Action & Exact & Repeated edit & Other pair & Unscorable \\
\midrule
\endhead
\midrule
\endfoot
\bottomrule
\endlastfoot
64 & Numerator & Generic & 0 & 72 & 0 & 0 \\
64 & Numerator & Source & 0 & 72 & 0 & 0 \\
64 & Denominator & Generic & 0 & 70 & 2 & 0 \\
64 & Denominator & Source & 0 & 72 & 0 & 0 \\
64 & Both & Generic & 0 & 70 & 2 & 0 \\
64 & Both & Source & 0 & 72 & 0 & 0 \\
192 & Numerator & Generic & 0 & 72 & 0 & 0 \\
192 & Numerator & Source & 0 & 72 & 0 & 0 \\
192 & Denominator & Generic & 0 & 70 & 2 & 0 \\
192 & Denominator & Source & 0 & 72 & 0 & 0 \\
192 & Both & Generic & 0 & 70 & 2 & 0 \\
192 & Both & Source & 0 & 72 & 0 & 0 \\
\end{longtable}\endgroup

%% file: ledger-budget-appendix.tex
\subsection{Gemma 4 31B}
\begingroup\small\setlength{\tabcolsep}{3.2pt}
\begin{longtable}{@{}llrrrrrrr@{}}
\caption{Gemma 4 31B ledger at 192 review tokens. Action entries show correct / valid out of 24. The Restart outcome is shared. Repeats reproduce a corrupted pair on the first line.}\label{tab:ledger-gemma31b-192} \\
\toprule
Operands & Proposal & Keep & Execute & Generic & Source & Restart & Generic repeats & Source repeats \\
\midrule
\endfirsthead
\toprule
Operands & Proposal & Keep & Execute & Generic & Source & Restart & Generic repeats & Source repeats \\
\midrule
\endhead
\midrule
\endfoot
\bottomrule
\endlastfoot
Exact & Correct & 24/24 & 24/24 & 23/24 & 24/24 & -- & 0 & 0 \\
Exact & Conflict & 0/24 & 24/24 & 21/24 & 20/24 & -- & 0 & 0 \\
Numerator & Correct & 24/24 & 0/24 & 4/24 & 10/24 & -- & 12 & 15 \\
Numerator & Conflict & 0/24 & 0/24 & 1/24 & 7/24 & -- & 15 & 16 \\
Numerator & Coherent & 0/24 & 0/24 & 0/24 & 5/24 & -- & 17 & 20 \\
Denominator & Correct & 24/24 & 0/24 & 3/24 & 3/24 & -- & 2 & 14 \\
Denominator & Conflict & 0/24 & 0/24 & 0/24 & 0/24 & -- & 6 & 12 \\
Denominator & Coherent & 0/24 & 0/24 & 1/24 & 1/24 & -- & 12 & 17 \\
Both & Correct & 24/24 & 0/24 & 2/24 & 3/24 & -- & 8 & 13 \\
Both & Conflict & 0/24 & 0/24 & 1/24 & 1/24 & -- & 5 & 12 \\
Both & Coherent & 0/24 & 0/24 & 0/24 & 1/24 & -- & 12 & 15 \\
Shared & -- & -- & -- & -- & -- & 2/24 & -- & -- \\
\end{longtable}\endgroup

\begingroup\small\setlength{\tabcolsep}{3.2pt}
\begin{longtable}{@{}llrrrrrrr@{}}
\caption{Gemma 4 31B ledger at 512 review tokens. Action entries show correct / valid out of 24. The Restart outcome is shared. Repeats reproduce a corrupted pair on the first line.}\label{tab:ledger-gemma31b-512} \\
\toprule
Operands & Proposal & Keep & Execute & Generic & Source & Restart & Generic repeats & Source repeats \\
\midrule
\endfirsthead
\toprule
Operands & Proposal & Keep & Execute & Generic & Source & Restart & Generic repeats & Source repeats \\
\midrule
\endhead
\midrule
\endfoot
\bottomrule
\endlastfoot
Exact & Correct & 24/24 & 24/24 & 24/24 & 24/24 & -- & 0 & 0 \\
Exact & Conflict & 0/24 & 24/24 & 24/24 & 24/24 & -- & 0 & 0 \\
Numerator & Correct & 24/24 & 0/24 & 24/24 & 24/24 & -- & 12 & 15 \\
Numerator & Conflict & 0/24 & 0/24 & 24/24 & 24/24 & -- & 15 & 16 \\
Numerator & Coherent & 0/24 & 0/24 & 24/24 & 24/24 & -- & 17 & 20 \\
Denominator & Correct & 24/24 & 0/24 & 24/24 & 24/24 & -- & 2 & 14 \\
Denominator & Conflict & 0/24 & 0/24 & 23/24 & 24/24 & -- & 6 & 12 \\
Denominator & Coherent & 0/24 & 0/24 & 24/24 & 24/24 & -- & 12 & 17 \\
Both & Correct & 24/24 & 0/24 & 24/24 & 24/24 & -- & 8 & 13 \\
Both & Conflict & 0/24 & 0/24 & 24/24 & 24/24 & -- & 5 & 12 \\
Both & Coherent & 0/24 & 0/24 & 22/24 & 24/24 & -- & 12 & 15 \\
Shared & -- & -- & -- & -- & -- & 22/24 & -- & -- \\
\end{longtable}\endgroup

\begingroup\small\setlength{\tabcolsep}{3.2pt}
\begin{longtable}{@{}llrrrrrrr@{}}
\caption{Gemma 4 31B ledger at 1024 review tokens. Action entries show correct / valid out of 24. The Restart outcome is shared. Repeats reproduce a corrupted pair on the first line.}\label{tab:ledger-gemma31b-1024} \\
\toprule
Operands & Proposal & Keep & Execute & Generic & Source & Restart & Generic repeats & Source repeats \\
\midrule
\endfirsthead
\toprule
Operands & Proposal & Keep & Execute & Generic & Source & Restart & Generic repeats & Source repeats \\
\midrule
\endhead
\midrule
\endfoot
\bottomrule
\endlastfoot
Exact & Correct & 24/24 & 24/24 & 24/24 & 24/24 & -- & 0 & 0 \\
Exact & Conflict & 0/24 & 24/24 & 24/24 & 24/24 & -- & 0 & 0 \\
Numerator & Correct & 24/24 & 0/24 & 24/24 & 24/24 & -- & 12 & 15 \\
Numerator & Conflict & 0/24 & 0/24 & 24/24 & 24/24 & -- & 15 & 16 \\
Numerator & Coherent & 0/24 & 0/24 & 24/24 & 24/24 & -- & 17 & 20 \\
Denominator & Correct & 24/24 & 0/24 & 24/24 & 24/24 & -- & 2 & 14 \\
Denominator & Conflict & 0/24 & 0/24 & 24/24 & 24/24 & -- & 6 & 12 \\
Denominator & Coherent & 0/24 & 0/24 & 24/24 & 24/24 & -- & 12 & 17 \\
Both & Correct & 24/24 & 0/24 & 24/24 & 24/24 & -- & 8 & 13 \\
Both & Conflict & 0/24 & 0/24 & 24/24 & 24/24 & -- & 5 & 12 \\
Both & Coherent & 0/24 & 0/24 & 24/24 & 24/24 & -- & 12 & 15 \\
Shared & -- & -- & -- & -- & -- & 24/24 & -- & -- \\
\end{longtable}\endgroup

\subsection{Qwen3-8B}
\begingroup\small\setlength{\tabcolsep}{3.2pt}
\begin{longtable}{@{}llrrrrrrr@{}}
\caption{Qwen3-8B ledger at 192 review tokens. Action entries show correct / valid out of 24. The Restart outcome is shared. Repeats reproduce a corrupted pair on the first line.}\label{tab:ledger-qwen8b-192} \\
\toprule
Operands & Proposal & Keep & Execute & Generic & Source & Restart & Generic repeats & Source repeats \\
\midrule
\endfirsthead
\toprule
Operands & Proposal & Keep & Execute & Generic & Source & Restart & Generic repeats & Source repeats \\
\midrule
\endhead
\midrule
\endfoot
\bottomrule
\endlastfoot
Exact & Correct & 24/24 & 24/24 & 24/24 & 24/24 & -- & 0 & 0 \\
Exact & Conflict & 0/24 & 24/24 & 23/24 & 24/24 & -- & 0 & 0 \\
Numerator & Correct & 24/24 & 0/24 & 0/24 & 0/24 & -- & 24 & 24 \\
Numerator & Conflict & 0/24 & 0/24 & 0/24 & 0/24 & -- & 24 & 24 \\
Numerator & Coherent & 0/24 & 0/24 & 0/24 & 0/24 & -- & 24 & 24 \\
Denominator & Correct & 24/24 & 0/24 & 1/24 & 0/24 & -- & 23 & 24 \\
Denominator & Conflict & 0/24 & 0/24 & 0/24 & 0/24 & -- & 23 & 24 \\
Denominator & Coherent & 0/24 & 0/24 & 0/24 & 0/24 & -- & 24 & 24 \\
Both & Correct & 24/24 & 0/24 & 0/24 & 0/24 & -- & 22 & 24 \\
Both & Conflict & 0/24 & 0/24 & 0/24 & 0/24 & -- & 24 & 24 \\
Both & Coherent & 0/24 & 0/24 & 0/24 & 0/24 & -- & 24 & 24 \\
Shared & -- & -- & -- & -- & -- & 0/1 & -- & -- \\
\end{longtable}\endgroup

\begingroup\small\setlength{\tabcolsep}{3.2pt}
\begin{longtable}{@{}llrrrrrrr@{}}
\caption{Qwen3-8B ledger at 512 review tokens. Action entries show correct / valid out of 24. The Restart outcome is shared. Repeats reproduce a corrupted pair on the first line.}\label{tab:ledger-qwen8b-512} \\
\toprule
Operands & Proposal & Keep & Execute & Generic & Source & Restart & Generic repeats & Source repeats \\
\midrule
\endfirsthead
\toprule
Operands & Proposal & Keep & Execute & Generic & Source & Restart & Generic repeats & Source repeats \\
\midrule
\endhead
\midrule
\endfoot
\bottomrule
\endlastfoot
Exact & Correct & 24/24 & 24/24 & 24/24 & 24/24 & -- & 0 & 0 \\
Exact & Conflict & 0/24 & 24/24 & 23/24 & 24/24 & -- & 0 & 0 \\
Numerator & Correct & 24/24 & 0/24 & 0/24 & 0/24 & -- & 24 & 24 \\
Numerator & Conflict & 0/24 & 0/24 & 0/24 & 0/24 & -- & 24 & 24 \\
Numerator & Coherent & 0/24 & 0/24 & 0/24 & 0/24 & -- & 24 & 24 \\
Denominator & Correct & 24/24 & 0/24 & 1/24 & 0/24 & -- & 23 & 24 \\
Denominator & Conflict & 0/24 & 0/24 & 0/24 & 0/24 & -- & 23 & 24 \\
Denominator & Coherent & 0/24 & 0/24 & 0/24 & 0/24 & -- & 24 & 24 \\
Both & Correct & 24/24 & 0/24 & 1/24 & 0/24 & -- & 22 & 24 \\
Both & Conflict & 0/24 & 0/24 & 0/24 & 0/24 & -- & 24 & 24 \\
Both & Coherent & 0/24 & 0/24 & 0/24 & 0/24 & -- & 24 & 24 \\
Shared & -- & -- & -- & -- & -- & 0/8 & -- & -- \\
\end{longtable}\endgroup

\begingroup\small\setlength{\tabcolsep}{3.2pt}
\begin{longtable}{@{}llrrrrrrr@{}}
\caption{Qwen3-8B ledger at 1024 review tokens. Action entries show correct / valid out of 24. The Restart outcome is shared. Repeats reproduce a corrupted pair on the first line.}\label{tab:ledger-qwen8b-1024} \\
\toprule
Operands & Proposal & Keep & Execute & Generic & Source & Restart & Generic repeats & Source repeats \\
\midrule
\endfirsthead
\toprule
Operands & Proposal & Keep & Execute & Generic & Source & Restart & Generic repeats & Source repeats \\
\midrule
\endhead
\midrule
\endfoot
\bottomrule
\endlastfoot
Exact & Correct & 24/24 & 24/24 & 24/24 & 24/24 & -- & 0 & 0 \\
Exact & Conflict & 0/24 & 24/24 & 23/24 & 24/24 & -- & 0 & 0 \\
Numerator & Correct & 24/24 & 0/24 & 0/24 & 0/24 & -- & 24 & 24 \\
Numerator & Conflict & 0/24 & 0/24 & 0/24 & 0/24 & -- & 24 & 24 \\
Numerator & Coherent & 0/24 & 0/24 & 0/24 & 0/24 & -- & 24 & 24 \\
Denominator & Correct & 24/24 & 0/24 & 1/24 & 0/24 & -- & 23 & 24 \\
Denominator & Conflict & 0/24 & 0/24 & 0/24 & 0/24 & -- & 23 & 24 \\
Denominator & Coherent & 0/24 & 0/24 & 0/24 & 0/24 & -- & 24 & 24 \\
Both & Correct & 24/24 & 0/24 & 0/24 & 0/24 & -- & 22 & 24 \\
Both & Conflict & 0/24 & 0/24 & 1/24 & 0/24 & -- & 24 & 24 \\
Both & Coherent & 0/24 & 0/24 & 0/24 & 0/24 & -- & 24 & 24 \\
Shared & -- & -- & -- & -- & -- & 15/23 & -- & -- \\
\end{longtable}\endgroup